\pdfoutput=1
\documentclass[pdflatex,sn-nature]{sn-jnl}%

\usepackage{graphicx}%
\usepackage{multirow}%
\usepackage{amsmath,amssymb,amsfonts}%
\usepackage{xcolor}%
\usepackage{textcomp}%
\usepackage{booktabs}%
\usepackage{float}%
\usepackage{threeparttable}%
\usepackage{pdfpages}

\setcitestyle{super}

\begin{document}

\title[Wavenumber-aware deep learning for Raman spectra]{Wavenumber-Aware Deep Learning for Robust Analysis of Raman Spectra}

\author[1]{\fnm{Jihwan} \sur{Kim}}
\equalcont{These authors contributed equally to this work.}

\author[2]{\fnm{Jaeseong} \sur{Heo}}
\equalcont{These authors contributed equally to this work.}

\author[2]{\fnm{Yujin} \sur{Kim}}

\author[3]{\fnm{Sang Bin} \sur{Yoon}}

\author[4]{\fnm{Chanjoong} \sur{Kim}}

\author[5]{\fnm{Kwang-Im} \sur{Oh}}

\author[3]{\fnm{Hyon Bin} \sur{Na}}

\author*[6,7]{\fnm{Kyungwon} \sur{Kwak}}\email{kkwak@korea.ac.kr}

\author*[2]{\fnm{Myeongkee} \sur{Park}}\email{bikeplay@pknu.ac.kr}

\author*[1]{\fnm{Joonyoung F.} \sur{Joung}}\email{jjoung@kookmin.ac.kr}

\affil[1]{\orgdiv{Department of Chemistry}, \orgname{Kookmin University}, \orgaddress{\street{77 Jeongneung-ro, Seongbuk-gu}, \city{Seoul}, \postcode{02707}, \country{Republic of Korea}}}

\affil[2]{\orgdiv{BB21+, Department of Chemistry}, \orgname{Pukyong National University}, \orgaddress{\city{Busan}, \postcode{48513}, \country{Republic of Korea}}}

\affil[3]{\orgdiv{Department of Chemical Engineering}, \orgname{Myongji University}, \orgaddress{\city{Yongin}, \postcode{17058}, \country{Republic of Korea}}}

\affil[4]{\orgdiv{School of Artificial Intelligence}, \orgname{Kookmin University}, \orgaddress{\street{77 Jeongneung-ro, Seongbuk-gu}, \city{Seoul}, \postcode{02707}, \country{Republic of Korea}}}

\affil[5]{\orgname{Institute of Basic Science, Korea University}, \orgaddress{\city{Seoul}, \postcode{02841}, \country{Republic of Korea}}}

\affil[6]{\orgdiv{Center for Molecular Spectroscopy and Dynamics}, \orgname{Institute for Basic Science (IBS), Korea University}, \orgaddress{\city{Seoul}, \postcode{02841}, \country{Republic of Korea}}}

\affil[7]{\orgdiv{Department of Chemistry}, \orgname{Korea University}, \orgaddress{\city{Seoul}, \postcode{02841}, \country{Republic of Korea}}}

\abstract{Machine-learning models for spectroscopy usually treat a spectrum as a fixed-grid intensity vector, leaving the physical coordinate implicit in the array index. Such models are tied to a single sampling grid, spectral range and resolution, which limits their use when measurement conditions change. Here we represent each spectral point as a pair of intensity and wavenumber and compute Transformer positional encodings from the wavenumber itself. On Raman spectra of three bacterial species and their synthetic mixtures, this representation remained robust to noise, sparser sampling and partial spectral coverage across convolutional, residual and Transformer networks. In controlled comparisons, wavenumber-based encoding outperformed index-based and learned encodings under partial observation and retained its performance on sampling grids never seen during training, where index-based encoding degraded. These findings were reproduced on a public 30-isolate bacterial benchmark. Explicit physical coordinates offer a simple route to robust spectral analysis under variable measurement conditions.}

\keywords{Raman spectroscopy, deep learning, wavenumber-aware representation, bacterial identification, robust spectral analysis}

\maketitle

\section{Introduction}

Machine learning has become a standard tool for the analysis of spectroscopic measurements across modalities, including infrared and vibrational spectroscopy,\cite{ye2020machine, berisha2019deep, lim2025comparison} nuclear magnetic resonance,\cite{li2021deep, hu2024accurate, yeo2025discrimination} mass spectrometry,\cite{yilmaz2024sequence, bushuiev2025self} optical absorption and emission spectroscopy,\cite{ghosh2019deep, joung2020experimental, joung2021deep} and X-ray diffraction,\cite{park2017classification, lee2020deep} and the field has recently been surveyed.\cite{guo2025artificial}
Despite the diversity of these techniques, they share a common structure: each measurement is defined over an explicit physical coordinate, such as wavenumber, chemical shift, mass-to-charge ratio, wavelength, or diffraction angle, and the recorded spectrum is a set of intensity values at positions along this coordinate axis.

In nearly all of these applications, however, spectra are supplied to learning models as fixed-grid intensity vectors, in which the physical coordinate is encoded only implicitly through the array index.\cite{acquarelli2017convolutional, liu2017deep, cui2018modern, zhang2019deepspectra, yang2019deep, ho2019rapid, berisha2019deep, lussier2020deep, li2021deep, koyun2024ramanformer, yu2024fragment, luo2025ssnet}
The consequences of this convention have long been recognized in chemometrics as the calibration-transfer problem: models built on one instrument degrade on another because sampling positions, resolution, and spectral range differ, and decades of standardization methods have been developed to compensate for these differences.\cite{workman2018review, feudale2002transfer}
Deep-learning approaches inherit the same limitation, and recent work has addressed it through domain adaptation between instruments.\cite{zhang2026novel}
The representation itself, in which measured intensities are detached from their physical coordinates, has received comparatively little attention.

This convention was established with convolutional neural networks (CNNs), the first deep architectures to be widely adopted for spectra, and it is shared across modalities.
In powder X-ray diffraction, Park et al.\ classified crystal symmetry from 10,001 intensity values recorded over a fixed range of diffraction angles, and described the diffraction pattern as being treated like a picture rather than as a set of peak positions and intensities.\cite{park2017classification}
Lee et al.\ identified the constituent phases of multiphase mixtures from synthetic patterns of 4,501 intensity values in the same manner.\cite{lee2020deep}
In nuclear magnetic resonance, Hu et al.\ interpolated each $^{1}$H spectrum onto a grid of 28,000 points between $-2$ and 12~ppm before passing it to a CNN, and binned the $^{13}$C chemical shifts into 80 fixed intervals.\cite{hu2024accurate}
CNNs for near-infrared, infrared, and Raman spectra follow the same design.\cite{acquarelli2017convolutional, cui2018modern, zhang2019deepspectra, yang2019deep, lussier2020deep}
For Raman-based bacterial identification, Ho et al.\ established a large and widely used public benchmark of approximately 60,000 spectra across 30 reference isolates and classified them with a deep residual CNN.\cite{ho2019rapid}
Notably, their architecture avoided pooling layers and instead used strided convolutions to preserve the exact positions of spectral peaks, an architectural choice reflecting that positional information along the spectral axis is essential for spectral discrimination.
However, the network in all of these models receives only the intensities.
The physical coordinate does not appear in the input, and position along the axis is conveyed solely by the array index.

Transformer architectures have more recently been adopted because self-attention can relate distant spectral features regardless of their distance along the axis, which convolutional filters with local receptive fields capture less efficiently.\cite{vaswani2017attention}
However, the input convention has been remained unchanged.
For structure elucidation from infrared spectra, Alberts et al.\ interpolated each spectrum to 400 points and supplied the integer-valued intensities as a token sequence, so that the position of a token in the sequence stands for its wavenumber.\cite{alberts2024leveraging}
Their simulated pretraining spectra began at 400~cm$^{-1}$, whereas their experimental spectra began at 450 or 550~cm$^{-1}$.
The missing low-wavenumber region was therefore filled with zeros.
In addition, a separate model was pretrained for each starting wavenumber.
Wu et al.\ increased the sampling to 3,200 uniformly spaced points and grouped them into fixed-length patches within a self-attention embedding layer, padding unmeasured regions with zeros.\cite{wu2025transformer}
In hyperspectral imaging, SpectralFormer embeds groups of neighboring spectral channels and assigns positions by channel order.\cite{hong2022spectralformer}
It was trained separately on data from three sensors that record 200, 103, and 144 spectral channels.
Transformer models for Raman spectra follow the same pattern: 
RamanFormer performs mixture-component quantification using fixed-length spectral patches,\cite{koyun2024ramanformer} 
the fragment-fusion Transformer classifies single-cell spectra using biologically motivated spectral fragments,\cite{yu2024fragment} 
and related Transformer designs have been explored for other Raman classification tasks.\cite{liu2023classification}
However, unlike natural language, spectra are continuous physical measurements rather than sequences of discrete symbols. 
Each measured intensity is inherently associated with a physical coordinate, and together they define the spectral observation. 
Because standard Transformers require tokenized inputs, existing spectral Transformers introduce an explicit discretization step, dividing spectra into resampled points, patches, fragments, or band groups that are subsequently treated as tokens whose positions are given by the token index.

Mass spectrometry is the exception.
A tandem mass spectrum is processed as a list of peaks, each a pair of mass-to-charge ratio ($m/z$) and intensity, which is a sparse set rather than a sampled curve.
Transformer models for mass spectra encode the coordinate directly: Casanovo encodes the $m/z$ value of each peak with a set of fixed sinusoidal functions and its intensity with a learned linear layer, and uses the peaks without discretizing the $m/z$ axis,\cite{yilmaz2024sequence} and DreaMS expands each $m/z$ value into Fourier features.\cite{bushuiev2025self}
Both models retain only the most intense peaks of a spectrum (up to 150 in Casanovo and 60 to 100 in DreaMS).
Bushuiev et al.\ further reported that replacing the continuous $m/z$ encoding with binned $m/z$ tokens decreases performance.\cite{bushuiev2025self}
Optical, magnetic resonance, and diffraction spectra differ from such peak lists.
They are densely sampled continuous signals in which band shapes, shoulders, overlapping bands, and the baseline between bands are all part of the measurement, and every sampled point is an observation.
Even a region without any band is informative, because the absence of a band at a given coordinate is itself a measured result.
For these signals the array index has served as a substitute for the coordinate, and coordinate-based encoding of the individual sampled points has, to our knowledge, not been examined.

Although the fixed-grid convention enables the use of standard architectures, it introduces several conceptual and practical limitations. 
First, the learned representation depends on design choices such as the resampling grid, patch lengths, fragment boundaries, and embedding strategies;\cite{dosovitskiy2020image} 
even when these choices are well motivated, for example fragment boundaries aligned with biochemically meaningful spectral regions,\cite{yu2024fragment} they remain anchored to fixed positions on a predetermined spectral grid.
Second, a fixed grid ties the model to a particular spectral resolution, sampling density, and spectral range,\cite{koyun2024ramanformer, yu2024fragment, alberts2024leveraging, wu2025transformer} so that spectra acquired under other settings must be interpolated, padded, or handled by a separately trained model.
Third, spectral regions that are unavailable due to instrumental limitations, fluorescence interference, detector saturation, or preprocessing procedures are not directly compatible with a fixed-grid representation. 
As a result, existing approaches implicitly assume that spectra are complete and uniformly sampled.
Real-world measurements often violate these assumptions.

When all spectra are acquired on the same instrument over the same spectral range, these limitations are of little practical consequence: array indices are then in one-to-one correspondence with physical coordinates, and a model can learn this correspondence implicitly without loss of information.
These limitations become important when spectra are acquired using different instruments or under varying experimental conditions. 

\begin{figure}[H]
    \centering
    \includegraphics[width=1\linewidth]{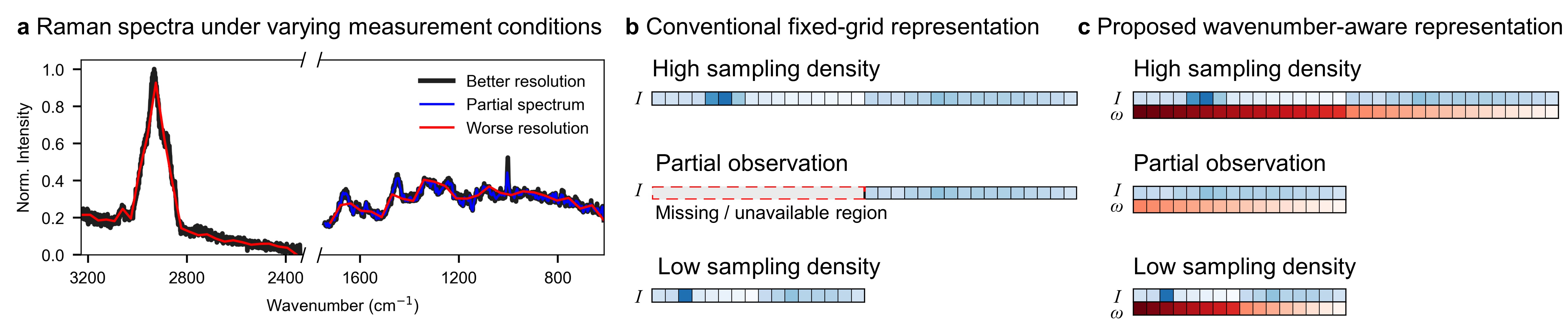}
    \caption{Overview of spectral variability and representation strategies for Raman data. 
    (a) Raman spectra acquired under varying measurement conditions, including differences in sampling density or resolution and partial observation due to missing or unreliable regions. 
    (b) Conventional fixed-grid representations encode spectra as one-dimensional signals with implicitly defined wavenumber coordinates, requiring a fixed discretization and full spectral coverage. Such representations are not well suited for spectra with varying sampling densities or incomplete observations. 
    (c) Proposed wavenumber-aware representation, where spectral intensity (\textit{I}) and corresponding wavenumber ($\omega$) are explicitly incorporated as separate input channels. This formulation enables flexible handling of varying sampling densities and partial spectral inputs without relying on a fixed discretization.}
    \label{overview}
\end{figure}

In this work, we examine this representational question in the context of Raman spectroscopy and bacterial identification, a setting in which its consequences are particularly acute.
Raman spectroscopy provides label-free vibrational fingerprints of molecular composition and has emerged as a promising route to rapid, reagent-free bacterial identification.\cite{butler2016using, kloss2013culture}
The task is demanding: bacterial cells share most of their biochemical components, so their spectra overlap strongly and the discriminative information is distributed across many vibrational bands rather than concentrated in isolated peaks.\cite{de2021biochemical, ho2019rapid}
Therefore, effective classification requires models capable of capturing long-range relationships among spectral features while remaining robust to measurement variability,\cite{koyun2024ramanformer, yu2024fragment, zhang2026novel} which makes bacterial identification a challenging task for the representational question raised above.
In practice, Raman spectra may differ in spectral resolution, sampling density, spectral range, and signal quality, often requiring interpolation, resampling, or model fine-tuning to compensate for instrument-specific differences.\cite{zhang2026novel} 
As illustrated in Fig.~\ref{overview}a, the same Raman signal may be acquired with different sampling densities, spectral ranges, or incomplete spectral coverage. 
Therefore, conventional fixed-grid representations require interpolation, resampling, or padding before inference, potentially introducing artifacts and reducing robustness across acquisition conditions.

We hypothesized that, under such varying acquisition conditions, the primary limitation lies not in the learning architecture itself but in the conventional fixed-grid representation of Raman spectra, where physical wavenumber information is implicitly encoded through sampling indices rather than explicitly represented.
A Raman spectrum is fundamentally a mapping between signal intensity and its corresponding physical coordinate (wavenumber), yet most existing approaches represent it solely as an intensity vector on a fixed sampling grid (Fig.~\ref{overview}b). 
Here, we explicitly incorporate wavenumber coordinates alongside intensity values, treating each spectral point as a physically defined observation rather than merely an index in a discretized sequence (Fig.~\ref{overview}c). 
In Transformer architectures, this corresponds to computing positional encodings directly from physical wavenumber coordinates rather than from sequence indices, enabling a controlled comparison against standard index-based and learned positional encodings under identical training conditions.
This representation naturally accommodates variations in spectral resolution, sampling density, and spectral coverage without requiring resampling onto a common grid at inference time.

In this study, we propose a wavenumber-aware deep learning approach for bacterial identification from Raman spectra. 
Using experimentally acquired spectra from multiple bacterial species, we evaluate the proposed representation under realistic measurement scenarios, including spectral mixing, additive noise, reduced sampling density, and partial spectral observation. 
We further benchmark wavenumber-aware inputs against constant, index-based, and learned positional encodings under identical data and augmentation conditions, and examine generalization to sampling grids whose wavenumber positions never appear during training.
The central comparison is carried out in parallel on our in-house dataset and on the independent 30-isolate public benchmark of Ho et al.\cite{ho2019rapid}
Our results demonstrate that explicitly incorporating physical wavenumber information provides a physically meaningful representation that improves robustness across diverse acquisition conditions. 
These properties are particularly relevant to practical Raman sensing scenarios, such as portable or field-deployed systems, in which spectra must be analyzed under variable and often degraded measurement conditions.

\section{Results and discussion}
\subsection{Realistic spectral variability and data augmentation}

\begin{figure}[H]
    \centering
    \includegraphics[width=1\linewidth]{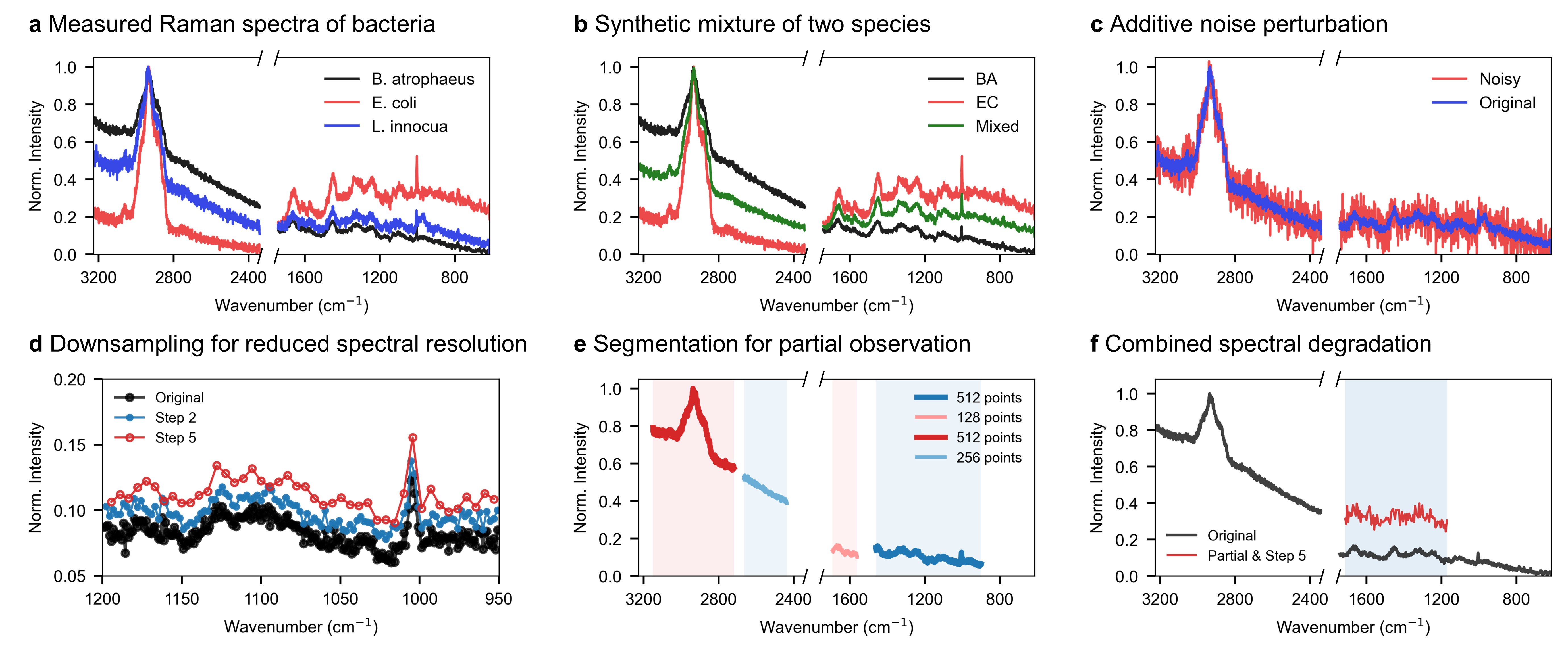}
    \caption{Data generation and augmentation pipeline for realistic Raman spectral modeling. 
    (a) Measured Raman spectra of three bacterial species, illustrating characteristic spectral features and the presence of missing regions. 
    (b) Synthetic mixture of two species generated by combining individual spectra with randomly sampled mixing ratios, enabling the modeling of spectral overlap. 
    (c) Additive Gaussian noise applied to simulate experimental variability and environmental perturbations. 
    (d) Downsampling of spectra to mimic reduced spectral resolution and sparse sampling conditions. 
    (e) Segmentation of spectral regions to represent partial observation, reflecting scenarios where only limited spectral windows are available. 
    (f) Combined spectral degradation incorporating downsampling, partial observation, and noise, simulating realistic sensing conditions encountered in field-deployed or portable Raman systems.}
    \label{data}
\end{figure}

Raman spectra acquired in practical environments rarely satisfy the assumptions underlying conventional fixed-grid representations.
Differences in instrumentation and acquisition conditions often lead to variations in spectral resolution, sampling density, spectral coverage, and signal quality, while portions of the spectrum may be unavailable because of detector limitations, fluorescence interference, saturation effects, or preprocessing procedures (Fig.~\ref{data}).

Importantly, this variability is not hypothetical: it is already present in the dataset used in this study, which consists of experimentally acquired Raman spectra of three bacterial species, \textit{Bacillus atrophaeus} (BA), \textit{Escherichia coli} (EC), and \textit{Listeria innocua} (LI).
The spectra were acquired under several different acquisition configurations.
Full-coverage spectra contain 2,048 points acquired in two disjoint spectral windows, a fingerprint window (614--1743~cm$^{-1}$, 1,024 points) and a high-wavenumber window (2341--3228~cm$^{-1}$, 1,024 points), separated by an unrecorded gap of approximately 600~cm$^{-1}$.
A subset of the LI spectra (400 out of 3,000) was acquired with the high-wavenumber window shifted to 2556--3415~cm$^{-1}$, and 2,000 out of the 2,500 BA spectra contain only the 1,024-point fingerprint window.
In addition, the sampling interval differs between windows (on average 1.10, 0.87, and 0.84~cm$^{-1}$ for the three window positions) and varies gradually within each window because of the nonlinear grating calibration.
As a result, there is no single index-to-wavenumber correspondence that holds across the dataset, and even spectra of the same species are not point-wise aligned.
A fixed-grid intensity vector cannot represent this structure without interpolation onto a common grid, whereas the explicit coordinate representation used in this work accommodates it directly.

All experiments derived from this dataset rest on three species measured in a single laboratory.
To test whether the conclusions transfer to an independently collected, substantially larger, and chemically more diverse dataset, we carried out the central experiments in parallel on the public bacterial Raman benchmark of Ho et al.\cite{ho2019rapid}
The benchmark comprises 60,000 reference spectra spanning 30 bacterial and yeast isolates, acquired on a different instrument, with 1,000 spectral points over 381.98--1792.4~cm$^{-1}$.
We applied the same multi-label formulation (synthetic mixtures of up to three isolates), the same augmentation strategy, and the same architectures, without task-specific tuning; the full protocol is described in the Methods section.
The in-house spectra contain heterogeneous native grids and the high-wavenumber region, whereas the benchmark spectra share a single fixed grid.

Because native spectral coverage partially correlates with species in the raw data, with fingerprint-only spectra occurring mainly for BA, care is required to prevent the coverage pattern from causing data leakage.
The windowing augmentation described below removes this risk: randomly positioned contiguous windows are applied to every class during training, so partial spectra occur for all three species and the extent of coverage carries no label information.
Full-coverage spectra are likewise available for all three species, including 500 BA spectra, so providing the full spectral range does not single out any class either.
In addition, all degradation evaluations impose identical windowing on every class.
Therefore, the augmentation strategy mirrors and extends variability that is already present in the measured data.

To emulate realistic measurement variability, we considered three common sources of spectral degradation: reduced sampling density, incomplete spectral coverage, and noise.
Reduced spectral resolution was simulated through stochastic downsampling of the original spectra (Fig.~\ref{data}d), whereas partial spectral observation was modeled by retaining only selected spectral regions (Fig.~\ref{data}e). 
Additive Gaussian noise was introduced to mimic experimental variability and reduced signal-to-noise ratios commonly encountered outside controlled laboratory environments (Fig.~\ref{data}c). 
To approximate realistic deployment scenarios, these perturbations were also applied simultaneously, resulting in spectra that exhibit reduced resolution, incomplete spectral coverage, and increased noise levels (Fig.~\ref{data}f). 
The magnitude of each perturbation was randomly sampled, resulting in spectra spanning a broad range of resolutions, spectral coverages, and noise levels.
Because downsampling is applied after windowing, the number of retained points is reduced by both operations in sequence: in principle, a 128-point window combined with a downsampling stride of 4 leaves only 32 measured points spanning roughly 140~cm$^{-1}$.
The joint severity of the combined perturbations was sampled randomly rather than explicitly controlled, so the combined scenarios can include cases substantially harsher than either perturbation alone, although we did not compile statistics of their joint distribution.

In addition, biological samples may contain multiple bacterial species whose Raman signatures overlap within a single spectrum. 
To simulate this situation, synthetic mixed spectra were generated by linearly combining experimentally measured spectra from different bacterial species using randomly sampled mixing ratios (Fig.~\ref{data}b). 
This procedure preserves experimentally observed spectral characteristics while generating diverse multi-species samples.
Synthetic mixtures were formulated as a multi-label classification problem, in which bacterial species contributing less than 10\% of the mixture were considered absent. 
Details are provided in the Methods section.
Together, these simulated perturbations and synthetic mixtures provide a realistic benchmark in which the ability of different spectral representations to accommodate variations in sampling density, spectral coverage, and measurement quality can be systematically evaluated.
Such conditions provide a demanding test of whether a spectral representation remains effective when the assumptions of conventional fixed-grid representations no longer hold.

\subsection{Evaluation across neural network architectures}

\begin{figure}[H]
    \centering
    \includegraphics[width=0.75\linewidth]{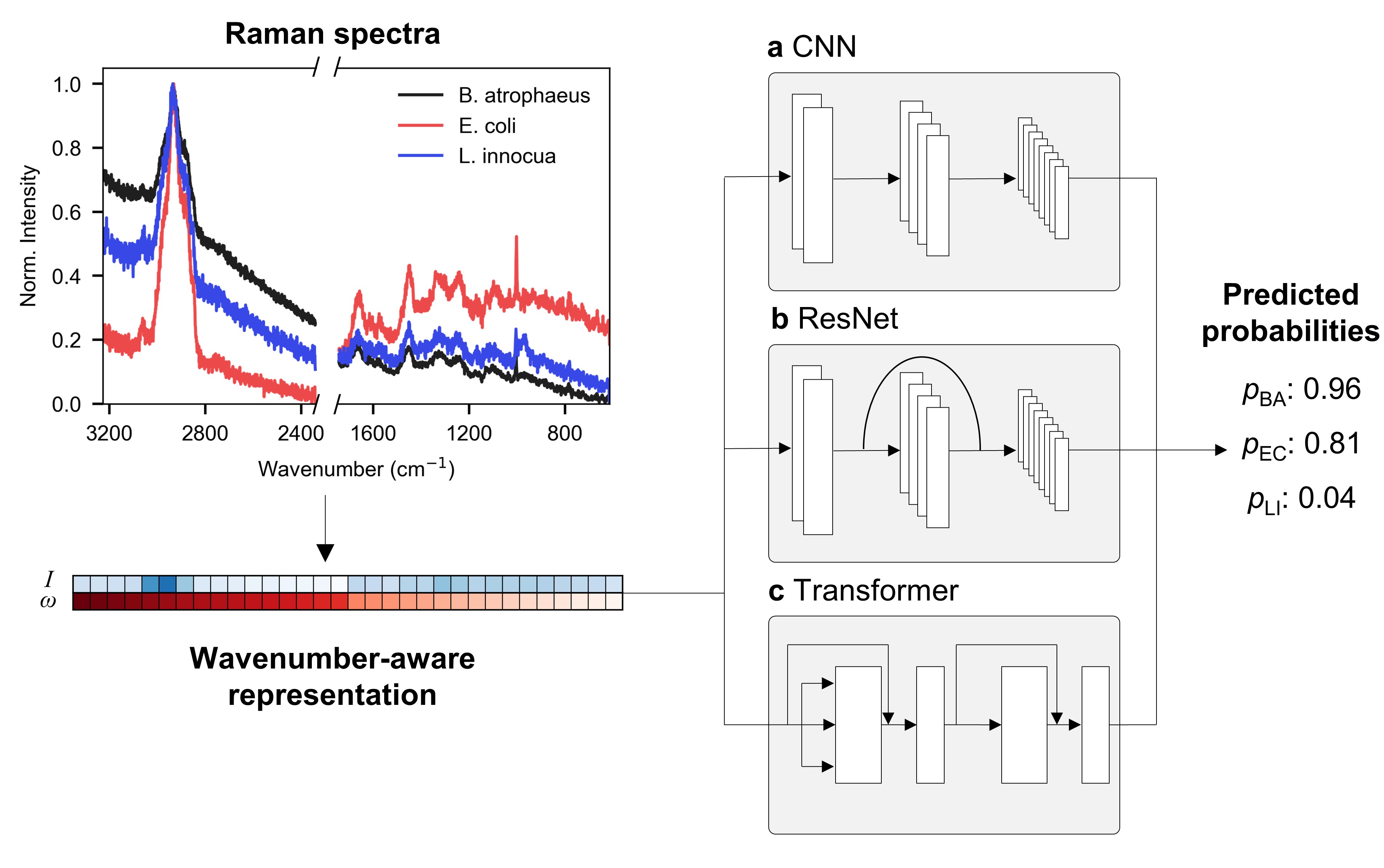}
    \caption{
    Comparison of neural network architectures operating on the proposed wavenumber-aware Raman representation.
    (a) CNN architecture for local spectral feature extraction using one-dimensional convolutions.
    (b) ResNet architecture employing residual blocks for hierarchical multi-scale spectral learning.
    (c) Transformer architecture treating each spectral point as an individual token and incorporating physical wavenumber information through positional encoding.
    All three architectures operate on the same wavenumber-aware representation, allowing differences in performance to be attributed to the learning architecture rather than the input representation.
    }
    \label{model}
\end{figure}

The overall model architectures used in this study are illustrated in Fig.~\ref{model}. 
To evaluate the proposed wavenumber-aware representation independently of a specific learning architecture, we employed three complementary neural network models: a convolutional neural network (CNN), a residual convolutional neural network (ResNet)\cite{he2016deep}, and a Transformer.\cite{vaswani2017attention} 
These architectures span representative families of deep learning models for spectral analysis, enabling us to assess whether the proposed representation consistently benefits fundamentally different learning paradigms. 
Detailed architectural descriptions and hyperparameter settings are provided in the Methods section and Supplementary Section~1.

For all architectures, each Raman spectrum was represented as paired measurements of normalized spectral intensity ($I$) and the corresponding physical wavenumber ($\omega$), forming a two-channel input representation. 
Spectra containing fewer sampled points because of downsampling or partial observation were padded to a fixed sequence length for efficient batch processing, while preserving the original spectral coordinates. 
Importantly, this padding strategy is used only for computational convenience and does not restrict the proposed representation to a particular spectral length or sampling scheme. 

All models were trained using the same input representation and identical multi-label classification objective.
Consequently, differences in predictive performance primarily reflect differences in learning architecture rather than differences in spectral representation.

\subsection{Robustness under realistic spectral degradation}

All models compared in this section were trained with the full augmentation strategy described above, in which noise, windowing, and downsampling were randomly applied during training.
Evaluation was then performed under five scenarios of increasing severity, starting from unperturbed spectra and adding the degradations one by one: Unperturbed, Noise only, Noise + Slice, Noise + Downsample, and Noise + Slice + Downsample.
Note that synthetic mixture generation is part of the task itself rather than a degradation: all evaluation sets, including the unperturbed scenario, consist of synthetic mixtures generated from the test split, so ``unperturbed'' refers only to the absence of noise, windowing, and downsampling.

The classification performances of the CNN, ResNet, and Transformer models on the in-house dataset under different spectral degradation scenarios are summarized in Table~\ref{tab:model_comparison}. 
Under the unperturbed condition, all three architectures achieved strong predictive performance, with F1$_{\mathrm{micro}}$ values exceeding 93\%. 
The three architectures performed comparably overall: the ResNet achieved the highest scores on unperturbed spectra (Exact Match 81.48\%, F1$_{\mathrm{micro}}$ 94.27\%), whereas the Transformer was the most robust under the scenarios involving partial spectral observation.
Architecture-level comparisons (CNN, ResNet, Transformer) on the public benchmark follow the same trends and are provided in Supplementary Tables~10 and 11.

\begin{table}[ht]
\centering
\caption{Performance comparison of CNN, ResNet, and Transformer models. The best performance under each scenario is highlighted in bold.}
\label{tab:model_comparison}
\begin{tabular}{lcccccc}
\toprule
& \multicolumn{2}{c}{CNN} & \multicolumn{2}{c}{ResNet} & \multicolumn{2}{c}{Transformer} \\
\cmidrule(lr){2-3} \cmidrule(lr){4-5} \cmidrule(lr){6-7}
Scenario
& Exact & F1$_{\mathrm{micro}}$ & Exact & F1$_{\mathrm{micro}}$ & Exact & F1$_{\mathrm{micro}}$ \\
\midrule
Unperturbed  & 79.95 & 93.68 & \textbf{81.48} & \textbf{94.27} & 80.02 & 93.78 \\
Noise only & 78.74 & 93.18 & \textbf{79.48} & \textbf{93.54} & 77.88 & 92.98 \\
Noise + Slice & 59.98 & 85.61 & 64.02 & 87.52 & \textbf{64.94} & \textbf{87.65} \\
Noise + Downsample & 73.82 & 91.33 & \textbf{74.75} & \textbf{91.78} & 73.59 & 91.31 \\
Noise + Slice + Downsample & 59.05 & 85.25 & 61.00 & 86.06 & \textbf{61.61} & \textbf{86.25} \\
\bottomrule
\end{tabular}
\end{table}

Because bacterial identification was formulated as a multi-label classification problem, Exact Match provides a strict metric that regards any mismatch in the predicted label set as incorrect. 
In contrast, F1$_{\mathrm{micro}}$ evaluates each bacterial label independently and therefore better reflects classification performance for mixed spectra.
Accordingly, F1$_{\mathrm{micro}}$ is used as the primary metric throughout the following discussion.

To evaluate robustness under realistic acquisition conditions, additive noise, reduced sampling density, partial spectral observation, and their combinations were introduced. 
Among these perturbations, additive Gaussian noise had only a minor effect on classification performance. 
Across all three architectures, F1$_{\mathrm{micro}}$ decreased by less than one percentage point, indicating that moderate experimental noise largely preserves the chemically informative vibrational signatures required for bacterial discrimination.

A similarly small degradation was observed under spectral downsampling. 
In this scenario the downsampling stride was drawn at random from 2 to 4 for each spectrum, so that between 50\% and 75\% of the original sampling points were removed. 
Under this condition, all models maintained F1$_{\mathrm{micro}}$ values above 91\%, with the highest value of 91.78\% obtained by the ResNet model, and the decrease relative to the unperturbed condition was approximately 2.5 percentage points for all architectures. 
Downsampling on the native grid has little effect once the models are trained with downsampling augmentation. 
Whether this robustness depends on the explicit wavenumber coordinate, and whether it extends to sampling grids that differ from the native grid, is examined in the following two sections.

In contrast, partial spectral observation resulted in substantially larger performance degradation. 
Under the Noise + Slice condition, F1$_{\mathrm{micro}}$ decreased by approximately 6--8 percentage points, whereas Exact Match accuracy decreased by 15--20 percentage points. 
Although the Transformer achieved the highest performance under this condition, the same trend was observed across all three architectures, indicating that the loss of spectral coverage is considerably more detrimental than a reduction in sampling density.

\begin{figure}[H]
    \centering
    \includegraphics[width=1\linewidth]{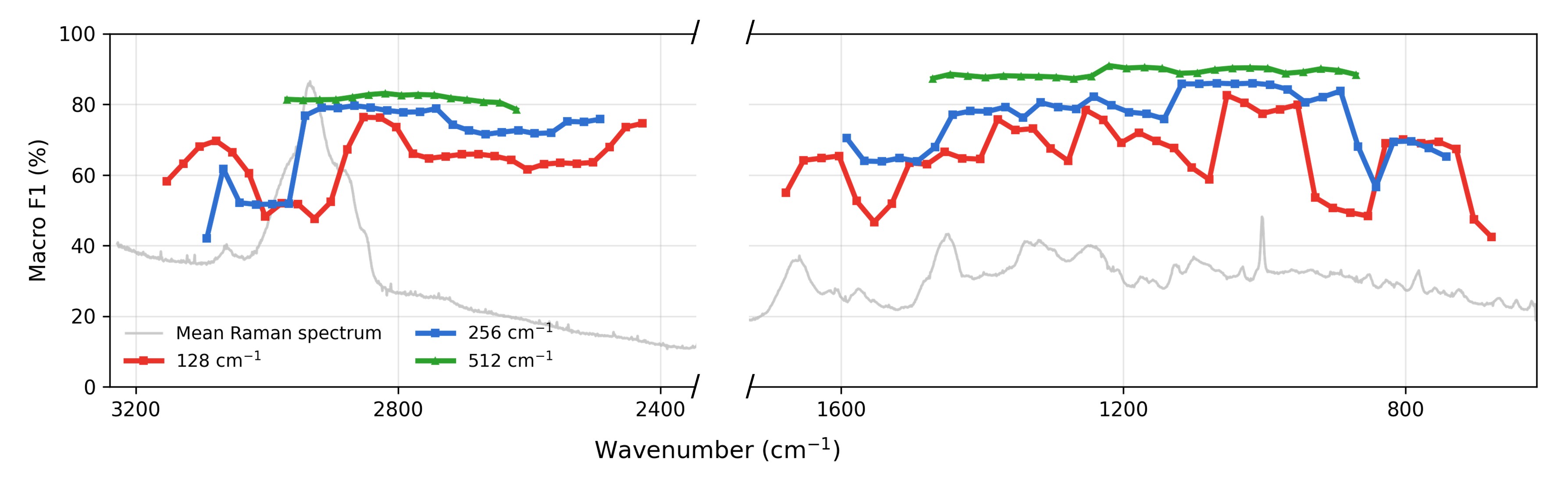}
    \caption{
    Sliding-window analysis of spectral-region dependence in Transformer-based classification.
    Macro F1 scores were evaluated using moving windows of 128, 256, and 512~cm$^{-1}$, plotted against the window center wavenumber.
    The gray trace represents the mean Raman spectrum averaged over all bacterial samples.
    }
    
    \label{sliding}
\end{figure}

To understand the origin of this performance degradation, we performed a sliding-window analysis using the Transformer model (Fig.~\ref{sliding}). 
Classification performance was evaluated using only the spectral information contained within a moving window, and the analysis was repeated for window widths of 128, 256, and 512~cm$^{-1}$.

The window width determined how strongly classification performance depended on window position.
With a 512~cm$^{-1}$ window, comparable to the minimum window length used during training (512 points, that is, roughly 560~cm$^{-1}$ in the fingerprint window and 430 to 450~cm$^{-1}$ in the high-wavenumber window), Macro F1 remained high at every window position, at about 87--91\% in the fingerprint window and 79--83\% in the high-wavenumber window.
A window of this width always contains several groups of vibrational bands.
Therefore, enough species-discriminative information is retained wherever the window is placed.

Narrower windows made performance increasingly dependent on window position.
The 256~cm$^{-1}$ window showed moderate fluctuations across the fingerprint region, with Macro F1 between about 57 and 86\%.
These fluctuations became much larger for the 128~cm$^{-1}$ window, which ranged from about 42 to 83\%.
A 128~cm$^{-1}$ window may cover only one or two groups of vibrational bands.
When these bands contain little species-discriminative information, the model has little basis for identifying the bacteria, and performance drops.

The C--H stretching region near 3000~cm$^{-1}$ deserves particular attention.
It contains the most intense band in the mean spectrum, yet both narrow windows performed poorly there.
Macro F1 fell to about 47--52\% for 128~cm$^{-1}$ windows centered between 2900 and 3030~cm$^{-1}$ and to 42--62\% for 256~cm$^{-1}$ windows centered above about 2970~cm$^{-1}$.
In both cases, the window covered the C--H stretching band but not the region below about 2850~cm$^{-1}$.
Once the window extended below this point, Macro F1 recovered to about 76--80\% for both widths.
Therefore, a high Raman intensity itself does not make a spectral region informative, because the C--H stretching band differs little between bacterial species.

These observations indicate that bacterial identification relies on chemically informative vibrational features rather than on overall Raman intensity, and that this holds even for windows narrower than any seen during training.
The fingerprint region contains numerous vibrational modes arising from proteins, lipids, nucleic acids, and carbohydrates, whose subtle spectral differences distinguish bacterial species.
In contrast, the C--H stretching region is dominated by chemically ubiquitous vibrational modes and therefore contributes less to species discrimination.
This observation also provides an independent cross-check of the fragment-based strategy of Yu et al.\cite{yu2024fragment}
Partitioning spectra into biologically meaningful regions, rather than weighting regions by raw intensity, targets the regions where the discriminative information actually lies.

Notably, introducing additional downsampling after spectral segmentation resulted in only marginal further degradation. 
For example, the Transformer achieved F1$_{\mathrm{micro}}$ scores of 87.65\% and 86.25\% under the Noise + Slice and Noise + Slice + Downsample conditions, respectively. 
Together, these results distinguish two fundamentally different forms of spectral degradation: reducing sampling density largely preserves chemically informative vibrational bands, whereas reducing spectral coverage removes them entirely.
Consequently, spectral coverage, rather than spectral resolution, emerges as the dominant factor limiting bacterial classification.

In short, additive noise and reduced sampling density had relatively minor effects on classification performance across all three architectures, whereas incomplete spectral coverage emerged as the primary source of performance degradation.
The class-wise predictions of the Transformer (Supplementary Fig.~1) show where this degradation occurs.
Errors were concentrated among mixtures containing multiple bacterial species, whereas single-species samples remained well distinguished even under combined spectral degradation.
Therefore, spectral degradation mainly increases ambiguity among compositionally similar mixtures rather than altering the learned class relationships.
All three architectures in this section receive the same wavenumber-aware input, so these results characterize the task and the degradations rather than the representation itself.
Whether the robustness originates from the physical coordinates is examined next.

\subsection{Comparison with standard positional encodings}

The preceding results raise a natural question: do the observed gains arise specifically from encoding \emph{physical wavenumber coordinates}, or could comparable robustness be obtained from data augmentation alone or from the positional encodings routinely used in standard Transformers?
To answer this question in a controlled manner, we trained Transformer models that differ \emph{only} in how positional information is supplied, keeping the data, synthetic-mixture generation, augmentation strategy, architecture, and training protocol identical.
Four encoding schemes were compared:
(i) \emph{constant} encoding, in which no positional information of any kind is provided;
(ii) \emph{index-based} sinusoidal encoding computed from sequence positions, corresponding to a standard Transformer;
(iii) \emph{learned} positional embeddings, which are trainable parameters optimized jointly with the model; and
(iv) the proposed \emph{wavenumber-based} sinusoidal encoding computed from physical coordinates.
Implementation details are provided in the Methods section, and the corresponding Exact Match results are reported in Supplementary Table~7.

\begin{table}[ht]
\centering
\caption{Comparison of positional-encoding schemes under identical data, augmentation, and training conditions on the in-house dataset and on the public benchmark of Ho et al.\ (Transformer, \%). All configurations use augmented training. For the in-house dataset, Exact Match results for all five scenarios are provided in Supplementary Table~7, and the remaining scenarios for the constant, index, and wavenumber encodings in Supplementary Table~9. The complete input-representation ablation on the benchmark is provided in Supplementary Table~14.}
\label{tab:pe_comparison}
\small\setlength{\tabcolsep}{2.5pt}
\begin{tabular}{llcccc}
\toprule
Dataset (metric) & Scenario &
Constant &
Index &
Learned &
Wavenumber \\
\midrule
\multirow{3}{*}{\shortstack[l]{In-house mixtures\\(F1$_{\mathrm{micro}}$)}}
 & Unperturbed            & 86.39 & 93.37 & 88.54 & \textbf{93.78} \\
 & Noise+Slice            & 77.72 & 81.69 & 76.93 & \textbf{87.65} \\
 & Noise+Slice+Downsample & 77.79 & 81.77 & 77.23 & \textbf{86.25} \\
\midrule
\multirow{3}{*}{\shortstack[l]{Benchmark single\\spectra (Top-1)}}
 & Unperturbed            & 27.28 & 93.96 & 92.54 & \textbf{95.90} \\
 & Noise+Slice            & 19.43 & 42.82 & 24.81 & \textbf{73.91} \\
 & Noise+Slice+Downsample & 18.51 & 49.30 & 45.19 & \textbf{58.36} \\
\midrule
\multirow{3}{*}{\shortstack[l]{Benchmark mixtures\\(Exact Match)}}
 & Unperturbed            &  6.19 & 52.57 & 49.22 & \textbf{57.17} \\
 & Noise+Slice            &  3.48 &  8.63 & 3.66 & \textbf{29.88} \\
 & Noise+Slice+Downsample &  2.83 & 13.64 & 11.70 & \textbf{19.14} \\
\bottomrule
\end{tabular}
\end{table}

Table~\ref{tab:pe_comparison} summarizes the results.
On unperturbed in-house spectra, index-based encoding is nearly equivalent to the wavenumber-based encoding (93.37\% versus 93.78\%), which is expected: although the native grids in our dataset are not exactly identical, they are similar enough that sequence indices remain in approximate one-to-one correspondence with physical wavenumbers, so either encoding conveys essentially the same information.
Learned positional embeddings remained below both sinusoidal encodings on unperturbed spectra (88.54\%), indicating that positional structure inferred from data alone is less effective than a fixed sinusoidal map for this task.
When spectra are partially observed (Noise+Slice), the gap between wavenumber-based and index-based encoding widens to approximately 6 percentage points (87.65\% versus 81.69\%), and learned embeddings fall below even the no-position baseline (76.93\% versus 77.72\%).
The mechanism is straightforward: after a spectral window is extracted, the retained points are re-indexed from zero, so index-based encoding conveys only the position \emph{within} the observed window and loses all information about \emph{where} in the spectrum that window lies.
Physical wavenumber coordinates, in contrast, are unaffected by windowing, and the model retains the absolute spectral location of every observed point.
We note that supplying absolute grid indices instead would not resolve this limitation, because recovering absolute indices presupposes knowledge of the acquisition grid, which is precisely the fixed-grid assumption the proposed representation removes.

In contrast, under downsampling alone, the two sinusoidal encodings remain close (91.31\% versus 90.72\% under Noise+Downsample, Supplementary Table~9), and even the constant encoding reaches 85.15\%.
When every $k$-th point of the native grid is retained, the index-to-wavenumber relation remains linear; only its slope changes, and the training augmentation exposes the model to every slope from 1 to 4.
Moreover, the slope of a given spectrum can be inferred from its length, so the index still identifies the physical coordinate almost completely.
Therefore, the benefit of the physical coordinate is specific to conditions in which the correspondence between index and wavenumber is broken rather than merely rescaled: under windowing, where the index restarts at an unknown wavenumber, and on sampling grids that differ from the native grid as shown in the next section.

The public benchmark shows the same ordering with a distinct contrast (Table~\ref{tab:pe_comparison}).
On unperturbed single spectra, index-based and wavenumber-based encodings perform comparably (93.96\% versus 95.90\% Top-1), as expected for spectra that share a single fixed grid.
Under Noise+Slice, however, the wavenumber-based encoding reaches 73.91\% Top-1, whereas the index-based encoding falls to 42.82\%.
This 31-percentage-point gap on a 30-class task is about five times larger than the corresponding gap on the in-house dataset.
Learned embeddings come close to the sinusoidal encodings on unperturbed benchmark spectra (92.54\%) but degrade most under partial observation.
Under Noise+Slice, they reach only 24.81\%, closer to the no-position baseline (19.43\%) than to the index-based encoding.
Synthetic mixtures of up to three isolates follow the same ordering.
Under the published evaluation protocol of Ho et al., the wavenumber-aware Transformer reaches 83.00\% Top-1 and 96.47\% group accuracy, comparable to the reported 82.2$\pm$0.3\% and 97.0$\pm$0.3\% (Supplementary Section~5 and Supplementary Table~12).\cite{ho2019rapid}
This protocol involves an exposure-time shift from 1-s training spectra to 2-s test spectra, and the corresponding zero-shot analysis is given in Supplementary Table~13.

The same experimental design also separates the roles of wavenumber information and data augmentation.
Without augmentation, adding the wavenumber channel to the intensity-only input raises F1$_{\mathrm{micro}}$ on unperturbed spectra from 84.75\% to 95.71\% (Table~\ref{tab:aug_ablation}), although part of this gain may reflect the coverage cue described in data augmentation section.
Augmentation alone yields a smaller change on unperturbed spectra (86.39\%).
Under degraded conditions, however, the wavenumber channel alone provides no robustness.
Without augmentation, the wavenumber-aware model falls to the level of the intensity-only model under Noise+Slice (61.70\% versus 62.14\%), because it has never been trained on incomplete inputs.
The benchmark shows the same pattern: without augmentation, the wavenumber-aware model reaches 94.30\% Top-1 on unperturbed spectra but only 17.14\% under Noise+Slice (Supplementary Table~14).
Conversely, augmentation without wavenumber information improves robustness but plateaus at 77.72\% under Noise+Slice.
Only the combined configuration attains high robustness (87.65\%).
It retains an 8--10 percentage-point advantage over the intensity-only model trained with identical augmentation (Tables~\ref{tab:pe_comparison} and \ref{tab:aug_ablation}).
Complete results for all three architectures and both metrics are given in Supplementary Tables~2--6.
Because these paired models differ only in the input representation, the remaining gain arises from explicitly preserving the physical correspondence between Raman intensity and wavenumber.
Therefore, data augmentation and the wavenumber-aware representation are complementary rather than interchangeable.
The explicit coordinates supply positional information that survives windowing.
However, the model learns to use this information only when augmentation exposes such degraded conditions during training.

\begin{table}[ht]
\centering
\caption{Effect of wavenumber information and data augmentation on Transformer performance (F1$_{\mathrm{micro}}$, \%). $I$ denotes the intensity-only input and $\omega$ the explicit wavenumber channel.}
\label{tab:aug_ablation}
\begin{tabular}{lcccc}
\toprule
Scenario & $I$ only & $I+\omega$ & $I$+aug.\tnote{a} & $I+\omega$+aug. \\
\midrule
Unperturbed & 84.75 & 95.71 & 86.39 & 93.78 \\
Noise+Slice & 62.14 & 61.70 & 77.72 & 87.65 \\
Noise+Slice+Downsample & 64.97 & 61.71 & 77.79 & 86.25 \\
\bottomrule
\end{tabular}
\begin{tablenotes}\footnotesize
\item[a] Identical to the Constant encoding in Table~\ref{tab:pe_comparison};
providing no positional information is equivalent to supplying a constant
coordinate channel.
\end{tablenotes}
\end{table}

\subsection{Generalization beyond the training distribution}

The preceding experiments demonstrated that the proposed wavenumber-aware representation provides robust performance under the spectral degradation conditions included during training. 
However, practical Raman measurements may exhibit substantially more severe variations in spectral coverage, sampling density, and noise than those represented in the training augmentation, for example, when acquisition settings or instrument configurations change.
To evaluate whether the proposed representation generalizes beyond the training distribution rather than simply learning the applied augmentation, we systematically evaluated the models trained on the in-house dataset under degradation levels exceeding those encountered during training.

\begin{figure}[H]
    \centering
    \includegraphics[width=1\linewidth]{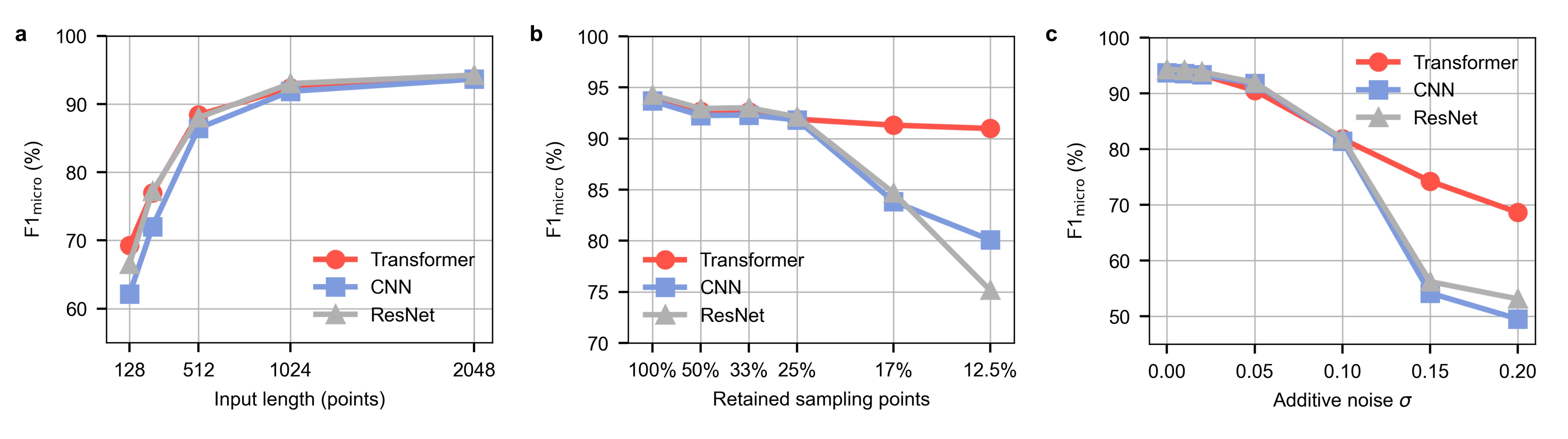}
    \caption{
    Generalization beyond the training distribution under increasingly severe spectral degradation.
    (a) Effect of reducing the available spectral length from 2048 points to 128 points (roughly 140~cm$^{-1}$). During training, spectra were segmented to lengths between 512 and 2048 points; shorter lengths were evaluated only during testing.
    (b) Effect of increasing the downsampling factor beyond the training range. The \textit{x}-axis indicates the percentage of retained sampling points, where 100\%, 50\%, 33\%, 25\%, 17\%, and 12.5\% correspond to downsampling factors of 1, 2, 3, 4, 6, and 8, respectively. Training employed downsampling factors between 1 and 4.
    (c) Effect of additive Gaussian noise with standard deviation $\sigma$. Training used noise levels within $\sigma \le 0.05$, whereas larger noise levels were evaluated only during testing.
    Performance is reported as F1$_{\mathrm{micro}}$ (\%).
    }
    \label{robustness}
\end{figure}

Figure~\ref{robustness} summarizes model performance under three types of distribution shift: progressively reduced spectral coverage, increasingly sparse spectral sampling, and additive Gaussian noise substantially larger than the augmentation range used during training.

Figure~\ref{robustness}a evaluates generalization to spectral windows shorter than those encountered during training. 
During training, spectra were segmented to lengths between 512 and 2048 points (roughly 560~cm$^{-1}$ and above), whereas 256- and 128-point windows (roughly 280 and 140~cm$^{-1}$) were evaluated only during testing. 
Consistent with the sliding-window analysis (Fig.~\ref{sliding}), all three architectures retained F1$_{\mathrm{micro}}$ above 86\% when at least 512 spectral points were available.
Below this range, performance declined noticeably to 72.0--77.2\% at 256 points and 62.1--69.2\% at 128 points. 
These results further support the conclusion that reliable bacterial discrimination depends on retaining sufficiently broad spectral coverage to capture chemically informative vibrational bands distributed across the Raman spectrum. 
Once the spectral window becomes too narrow, essential vibrational information is inevitably lost, limiting classification performance regardless of model architecture.

A noticeably different behavior was observed for spectral downsampling (Fig.~\ref{robustness}b).
Although the models were trained using downsampling factors of only 1--4, evaluation was extended to factors of 6 and 8, corresponding to retaining only 16.7\% and 12.5\% of the original sampling points, respectively. 
Within the training range, the three architectures behaved almost identically, with F1$_{\mathrm{micro}}$ of 91.8--92.1\% at a factor of 4.
Beyond this range, their behavior diverged.
The Transformer maintained F1$_{\mathrm{micro}}$ above 90\% at factors of 6 and 8 (91.31\% and 90.99\%), whereas the CNN and ResNet fell to 83.84\% and 84.68\% at a factor of 6 and to 80.09\% and 75.17\% at a factor of 8.
All three architectures receive the same wavenumber-aware input, so this difference reflects how each architecture uses the coordinates.
A convolution kernel spans a fixed number of points, so its physical width grows with the downsampling factor and eventually exceeds the range seen during training.
In the Transformer, each point carries its own coordinate through the positional encoding, and attention is not tied to a fixed number of neighboring points.
Therefore, as long as chemically informative vibrational bands remain sampled, the Transformer can associate spectral features with their correct spectral coordinates despite substantial reductions in sampling density.

Figure~\ref{robustness}c evaluates robustness against additive Gaussian noise beyond the augmentation range used during training. 
Within the training range ($\sigma \le 0.05$), all three architectures showed only minor degradation, retaining F1$_{\mathrm{micro}}$ of 90.5--91.9\% at $\sigma=0.05$.
At $\sigma=0.10$, all three declined to a similar level (81.4--81.9\%).
The architectures diverged only at larger noise levels.
At $\sigma=0.15$, the CNN and ResNet fell to 54.13\% and 56.18\%, whereas the Transformer retained 74.20\%.
At $\sigma=0.20$, four times the maximum noise level used during training, the Transformer still reached 68.60\%, compared with 49.49\% and 53.08\% for the CNN and ResNet.
Because all three architectures receive the same coordinates, this difference reflects the architecture rather than the input representation.
Convolutional models build their features from local patterns formed by neighboring points, and strong noise distorts these local band shapes.
In contrast, self-attention can combine evidence from bands distributed across the whole spectrum.
Therefore, even when noise obscures the shape of individual bands, the Transformer may still identify the bacteria from the discriminative bands that remain recognizable.

Beyond degradation severity, we also evaluated extrapolation along a second, qualitatively different axis: generalization to wavenumber coordinates that never appear during training.
All degradation experiments above retain subsets of the native measurement grid, so every evaluated $\omega$ value lies within the training distribution.
A model deployed on a different instrument, however, would encounter a fundamentally different sampling grid, for example a uniform 2~cm$^{-1}$ grid producing values such as 616, 618, and 620~cm$^{-1}$ that were never observed during training.
To test this scenario directly, evaluation spectra were resampled onto uniform wavenumber grids with spacings of 1, 2, and 4~cm$^{-1}$, deliberately misaligned with the native, nonuniform measurement grid, so that essentially none of the resulting coordinate values coincide with those seen during training (Table~\ref{tab:offset_grid}).

\begin{table}[ht]
\centering
\caption{Generalization to sampling grids whose wavenumber values never appear during training (Transformer, synthetic mixtures, F1$_{\mathrm{micro}}$, \%). Evaluation spectra were resampled onto uniform grids with the indicated spacing; ``native'' denotes the original measurement grid (in-house: nonuniform, 0.84--1.10~cm$^{-1}$; benchmark: uniform, about 1.41~cm$^{-1}$). Complete results are provided in Supplementary Tables~9 and 15.}
\label{tab:offset_grid}
\begin{tabular}{llcccc}
\toprule
Scenario & Encoding & Native & 1~cm$^{-1}$ & 2~cm$^{-1}$ & 4~cm$^{-1}$ \\
\midrule
\multicolumn{6}{l}{\emph{In-house dataset}} \\
\multirow{3}{*}{Unperturbed}
 & Wavenumber (proposed) & 93.78 & \textbf{93.74} & \textbf{93.68} & \textbf{92.44} \\
 & Index                 & 93.37 & 72.50 & 78.38 & 82.37 \\
 & Constant              & 86.39 & 85.80 & 85.64 & 84.80 \\
\multirow{3}{*}{Noise+Slice}
 & Wavenumber (proposed) & 87.65 & \textbf{88.01} & \textbf{87.47} & \textbf{86.31} \\
 & Index                 & 81.69 & 79.37 & 79.80 & 81.37 \\
 & Constant              & 77.72 & 77.84 & 78.09 & 77.79 \\
\midrule
\multicolumn{6}{l}{\emph{Benchmark of Ho et al.}} \\
\multirow{2}{*}{Unperturbed}
 & Wavenumber (proposed) & 80.08 & \textbf{73.22}\tnote{a} & \textbf{78.97} & \textbf{73.06} \\
 & Index                 & 76.63 & 7.51\tnote{a} & 23.35 & 24.86 \\
\multirow{2}{*}{Noise+Slice}
 & Wavenumber (proposed) & 51.69 & \textbf{51.89} & \textbf{49.35} & \textbf{40.21} \\
 & Index                 & 20.04 & 13.24 & 30.94 & 30.87 \\
\bottomrule
\end{tabular}
\begin{tablenotes}\footnotesize
\item[a] The resampled spectrum (about 1{,}410 points) exceeds the fixed input length of 1{,}024 points and was truncated at the high-wavenumber end.
\end{tablenotes}
\end{table}

A concern raised by this scenario is that although sinusoidal encodings can be \emph{computed} for arbitrary $\omega$ values, the attention mechanism might lack a learned basis for interpreting coordinates it has never encountered.
However, the results indicate otherwise: wavenumber-based encoding retains 92.4--93.7\% F1$_{\mathrm{micro}}$ across all unseen grids, essentially matching native-grid performance, because the sinusoidal map is continuous in $\omega$, so nearby coordinates receive nearly identical embeddings regardless of whether they appeared during training.
Index-based encoding, by contrast, collapses to as low as 72.5\% on the same inputs, since resampling alters both the number and the correspondence of sequence positions.
The residual decrease of the wavenumber-based encoding on coarser grids has a simple interpretation: the 1~cm$^{-1}$ grid provides a point density similar to the native grid, whereas the 2 and 4~cm$^{-1}$ grids also reduce the number of measured points, an effect analogous to downsampling.
Consistent with this interpretation, the small residual decrease on the 4~cm$^{-1}$ grid (93.78\% to 92.44\%) parallels the effect of downsampling applied on the native grid (Supplementary Table~9).
Complete results for all five evaluation scenarios are provided in Supplementary Table~9.
Notably, index-based encoding falls below even the constant baseline on the finer grids, indicating that misaligned positional information is worse than none at all.
Explicit physical coordinates are thus not merely an alternative way of indexing the spectrum; they are what makes generalization across sampling grids possible at all.

The benchmark shows the same contrast more (Table~\ref{tab:offset_grid}).
Its spectra share a single native grid with a spacing of about 1.41~cm$^{-1}$.
Therefore, the 1~cm$^{-1}$ grid is denser than any spectrum seen during training, because the augmentation only removes points.
On this grid, only the wavenumber-based encoding remained useful.
Under Noise+Slice, it reached 51.89\%, matching the native-grid value (51.69\%), whereas the index-based encoding fell from 20.04\% to 13.24\%.
On unperturbed spectra, the 1~cm$^{-1}$ grid exceeds the input length and the spectra were truncated.
Even so, the wavenumber-based encoding retained 73.22\%, whereas the index-based encoding dropped to 7.51\%.
On the 2 and 4~cm$^{-1}$ grids, the wavenumber-based encoding retained 78.97\% and 73.06\%, compared with 80.08\% on the native grid.
As in the in-house data, the decrease on the 4~cm$^{-1}$ grid parallels downsampling on the native grid (68.13\% under Noise+Downsample, Supplementary Table~15).
The index-based encoding collapsed to 23.35\% and 24.86\% on the same grids.
Once the grid changes, the one-to-one correspondence between index and wavenumber is lost, and the same collapse occurs even on a 1.5~cm$^{-1}$ grid close to the native spacing (18.90\%, Supplementary Table~15).
This collapse is far larger than on the in-house dataset, likely because every benchmark spectrum shares one grid, whereas the in-house native grids already differ between acquisition settings.

These experiments reveal two different limits of robust Raman classification.
The first limit is spectral coverage.
A window that is too narrow no longer contains the chemically informative bands.
No architecture can recover this missing information.
The second limit is the sampling of the spectrum, and it proved far less restrictive.
The Transformer was little affected by sparser sampling.
On sampling grids never seen during training, the wavenumber-based encoding retained most of its performance, while the index-based encoding did not.
Both findings go beyond the training distribution.
The degradation experiments (Fig.~\ref{robustness}) used coverage, sampling density, and noise levels more severe than those in the training augmentation.
The resampled-grid experiments (Table~\ref{tab:offset_grid}) used wavenumber coordinates that never appeared during training.
Therefore, the proposed representation captures physically meaningful spectral relationships rather than merely reproducing the augmentation distribution.

\subsection{Relation to architecture-level and domain-adaptation approaches}

The proposed representation is complementary to, rather than competitive with, recent architecture-level innovations for Raman Transformers.
Patch-based tokenization (RamanFormer\cite{koyun2024ramanformer}) and biologically motivated spectral fragmentation (fragment-fusion Transformer\cite{yu2024fragment}) improve feature extraction \emph{within} a fixed-grid representation, and were developed for tasks (chemical mixture-ratio quantification and single-cell phenotype classification, respectively) that differ from the bacterial identification problem studied here, precluding a meaningful direct numerical comparison.
Instead, our ablations isolate the representational question these designs leave open: wavenumber-aware point-wise tokens are compared against index-based and learned positional encodings under identical data, augmentation, and training conditions (Table~\ref{tab:pe_comparison}), which is the controlled comparison relevant to the claims of this work.
In principle, wavenumber-aware coordinates could also be injected into patch- or fragment-level tokens, combining both directions.

A similar complementarity holds for domain adaptation.
Zhang et al.\ recently showed that unsupervised domain adaptation substantially mitigates batch-, strain-, and instrument-induced shifts in Raman-based bacterial identification, with accuracy collapsing to 20.8\% under direct cross-instrument transfer without adaptation.\cite{zhang2026novel}
Domain adaptation aligns feature \emph{distributions} but presupposes fixed-length inputs; spectra from their second instrument were linearly interpolated onto the 1024-point grid of the first to satisfy the CNN backbone.
The representation proposed here removes precisely this input-stage constraint and could therefore serve as a natural front end for domain-adaptive training: explicit coordinates address grid heterogeneity at the input level, while adaptation addresses residual differences in spectral response that coordinates alone cannot capture.
Consistent with this division of labor, we do not claim that wavenumber-aware inputs enable zero-shot cross-instrument transfer; the exposure-shift experiment on the Ho et al.\ benchmark and the results of Zhang et al.\ both indicate that instrument-induced shifts extend well beyond coordinate mismatch.\cite{ho2019rapid, zhang2026novel}
Finally, we note that the proposed representation presupposes a calibrated wavenumber axis, which is standard practice in Raman instrumentation; robustness to residual calibration drift could be further promoted through wavenumber-jitter augmentation, which we leave for future work.

\section{Conclusion}

In this study, we introduced a wavenumber-aware representation for Raman spectral analysis in which spectral intensity and physical wavenumber coordinates are explicitly provided as separate input channels. 
Whereas conventional fixed-grid representations implicitly encode spectral coordinates through index positions, an encoding that is sufficient only as long as the sampling grid remains fixed, the proposed formulation directly preserves the physical relationship between Raman intensity and wavenumber, enabling flexible handling of spectra acquired under varying sampling densities, spectral ranges, and measurement conditions.

Using experimentally acquired bacterial Raman spectra, we systematically evaluated the proposed representation across CNN, ResNet, and Transformer architectures. 
The representation consistently improved classification performance and remained robust under realistic spectral degradation, including additive noise, reduced sampling density, and partial spectral observation.
Sliding-window analysis further demonstrated that bacterial discrimination depends primarily on retaining chemically informative vibrational bands rather than simply increasing the number of sampled spectral points. 
In addition, the proposed representation generalized to degradation levels and sampling grids beyond those encountered during training, indicating that the learned relationships extend beyond the augmentation distribution.
These conclusions were reproduced on the independent, tenfold-larger public benchmark of Ho et al.,\cite{ho2019rapid} where the proposed representation matched the published CNN baseline under its evaluation protocol while providing markedly higher robustness under partial spectral observation.

This work also shows the value of treating physical spectral coordinates as an integral part of the input representation rather than information implicitly defined by a discretized sampling grid. 
Although demonstrated here using Raman spectroscopy, the same principle may extend to other measurements defined over physical coordinates, such as wavenumber, wavelength, frequency, chemical shift, or mass-to-charge ratio.
Indeed, coordinate-based encodings of spectral peaks have recently and independently emerged in mass spectrometry,\cite{yilmaz2024sequence} suggesting that treating physical coordinates as first-class model inputs may develop into a shared design principle for spectral machine learning; validating such extensions is left for future work. 
Within Raman spectroscopy, explicitly incorporating these coordinates into spectral representations provides a simple yet physically meaningful route to robustness under variable acquisition conditions.

We anticipate that physically informed spectral representations will facilitate the development of more robust machine-learning models for Raman sensing, particularly in applications involving varying acquisition protocols, sparse sampling, or incomplete spectral observations.

\section{Methods}\label{method}
\subsection{Bacterial sample preparation and Raman measurements}
For bacterial sample preparation, approximately 0.05~g of freeze-dried cells of \textit{E. coli}, \textit{B. atrophaeus}, and \textit{L. innocua} were each dispersed in 1~mL of deionized water to prepare separate suspensions.
Subsequently, 10~$\mu$L of each suspension was deposited onto a glass substrate and allowed to dry under ambient conditions ($\sim$23~$^\circ$C) for 20~min prior to Raman measurements.

Raman spectra were acquired using a micro-Raman spectrometer (LabRAM HR, Horiba) equipped with a 633~nm He--Ne laser and a 1800~grooves~mm$^{-1}$ diffraction grating. 
The laser beam was focused onto the sample through a 100$\times$ objective lens (Olympus, NA~=~0.9). 
Wavenumber calibration was performed using the characteristic Raman bands of polystyrene powder. 
The spectral resolution was approximately 1~cm$^{-1}$. 
The laser power at the sample surface was adjusted to 0.7~mW to minimize thermal damage to the samples. 
For each spectrum, an acquisition time of 1~s and 100 accumulations were used.

\subsection{Dataset construction and data splitting}

The dataset consisted of experimentally acquired Raman spectra from three bacterial species: \textit{Bacillus atrophaeus} (BA), \textit{Escherichia coli} (EC), and \textit{Listeria innocua} (LI). A total of 7,500 spectra were collected, comprising 2,500 BA spectra, 2,000 EC spectra, and 3,000 LI spectra.

Each spectrum was acquired under one of several acquisition configurations.
Full-coverage spectra comprise two disjoint windows recorded in a single file: a fingerprint window (614--1743~cm$^{-1}$, 1,024 points) and a high-wavenumber window (2341--3228~cm$^{-1}$, 1,024 points), separated by an unrecorded region.
All EC spectra and 2,600 LI spectra follow this configuration; the remaining 400 LI spectra were acquired with the high-wavenumber window positioned at 2556--3415~cm$^{-1}$.
For 2,000 of the 2,500 BA spectra, only the 1,024-point fingerprint window was recorded, so these spectra contain 1,024 points.
The average sampling interval is 1.10~cm$^{-1}$ in the fingerprint window and 0.87 or 0.84~cm$^{-1}$ in the two high-wavenumber window positions, and the interval varies gradually within each window owing to the nonlinear grating calibration, so no two acquisition configurations share a common index-to-wavenumber mapping.

To ensure unbiased model evaluation, the dataset was randomly divided into training, validation, and test sets using a stratified split while preserving the class distribution of each bacterial species. Specifically, 80\% of the spectra were assigned to the training set, 10\% to the validation set, and 10\% to the test set. This resulted in 6,000 training spectra, 750 validation spectra, and 750 test spectra.

The number of spectra assigned to each split is summarized in Supplementary Table~16. All synthetic mixtures and data augmentations were generated independently within each split to prevent information leakage between training and evaluation datasets.

\subsection{Synthetic mixture generation and spectral perturbation}

To emulate realistic multi-species samples, synthetic Raman mixtures were generated by combining spectra from one to three bacterial species. 
For each sample, the number of constituent species was randomly selected from \{1, 2, 3\}, and mixing coefficients were drawn from a Dirichlet distribution:

\begin{equation}
I_{\mathrm{mix}}(\omega)
=
\sum_{i=1}^{N}
x_i I_i(\omega),
\qquad
\sum_{i=1}^{N} x_i = 1,
\end{equation}

where $I_i(\omega)$ denotes the Raman spectrum of species $i$ and $x_i$ is its corresponding mixing coefficient. 
Prior to mixing, each individual spectrum was independently normalized to the range $[0,1]$. 
The resulting mixed spectrum was subsequently normalized again to ensure a consistent intensity scale across all samples.

Because the constituent spectra of a synthetic mixture were acquired on slightly different instrument-native wavenumber grids, forming their weighted sum requires evaluating the spectra at common wavenumber positions.
We emphasize that this alignment is a data-synthesis step required only for constructing synthetic mixtures and their composition labels; it is not a preprocessing requirement of the proposed representation or of the models.
For each mixture, the common grid was selected anew from the measured coordinates of the spectra being combined, rather than from a predefined reference axis.
Specifically, the overlapping wavenumber range shared by the selected spectra was identified, and the measured grid of the spectrum containing the fewest points within this overlap was adopted as the target grid.
The remaining spectra were linearly interpolated onto this coarsest measured grid before weighted summation; no artificial uniform wavenumber grid was created at any stage, and the resulting mixture therefore retains the nonuniform coordinates of an actual measurement.
Single-species spectra were never interpolated and enter the models on their original measured coordinates.
Likewise, no resampling is applied at the model-input stage: all spectra are presented to the models on their native wavenumber grids.

Multi-label targets were assigned according to the mixture composition. 
A bacterial species was considered present when its mixing coefficient exceeded 0.1. 
Thus, the target label vector was defined as

\begin{equation}
y_i =
\begin{cases}
1, & x_i \ge 0.1 \\
0, & x_i < 0.1
\end{cases}
\end{equation}

for each bacterial species $i$.

To evaluate robustness under realistic measurement conditions, three spectral perturbations were considered.

First, additive Gaussian noise was introduced to simulate detector noise and experimental variability:

\begin{equation}
\tilde I(\omega)
=
I(\omega)
+
\mathcal{N}(0,\sigma^2),
\end{equation}

where the noisy intensities were clipped to the range $[0,1]$.
During training, $\sigma$ was drawn uniformly from $[0, 0.05]$ for each spectrum, whereas $\sigma=0.025$ was used during evaluation.

Second, partial spectral observation was simulated by retaining only a contiguous spectral segment. 
A fixed-length window was randomly selected along the Raman axis, and only the spectral points within this window were provided to the model.
During training, the window length was 2048, 1024, or 512 points with probabilities of 0.4, 0.3, and 0.3; a 2048-point window retains the full spectrum.
During evaluation, the window length was 512 points.

Third, reduced spectral resolution was simulated through downsampling. 
After slicing, every $k$-th spectral point was retained, starting from a randomly chosen offset.
During training, $k$ was randomly selected from \{1, 2, 3, 4\}, whereas during evaluation it was selected from \{2, 3, 4\}. 
This procedure mimics spectra acquired at lower spectral resolution or under accelerated acquisition conditions.

Five evaluation scenarios were considered: (i) Unperturbed, (ii) Noise only, (iii) Noise + Slice, (iv) Noise + Downsample, and (v) Noise + Slice + Downsample. 
In scenario (v), the number of retained points is the product of the windowing and downsampling reductions, so this scenario includes spectra that are considerably sparser than in either scenario (iii) or (iv). 
For each scenario, 10,000 synthetic mixtures were generated from the test split and evaluated using a prediction threshold of 0.5 for all models.

\subsection{Machine learning model architectures}
\paragraph{CNN.}

The CNN model serves as a baseline architecture for Raman spectral classification.
The input spectrum is treated as a two-channel one-dimensional signal and processed through a series of one-dimensional convolutional layers. 
These convolutional filters learn local spectral patterns by aggregating information from neighboring spectral points. 
Pooling operations are subsequently applied to progressively reduce the spectral dimension while increasing the receptive field of the network. 
The resulting feature representation is aggregated and passed through a fully connected classification layer to produce multi-label predictions. 
Because CNNs primarily rely on local feature extraction, they provide a useful reference for evaluating whether the proposed representation remains effective under architectures with limited global context.

\paragraph{ResNet.}

The ResNet model extends the convolutional framework by incorporating residual connections between convolutional blocks. 
These skip connections facilitate gradient propagation during training and enable substantially deeper architectures without performance degradation. 
Similar to the CNN model, the input spectrum is processed as a two-channel one-dimensional signal.
As the network depth increases, the spectral dimension is progressively reduced through strided convolutions while the number of feature channels is increased, resulting in hierarchical representations that capture spectral features at multiple scales. 
The final feature map is aggregated using global average pooling and subsequently passed to a fully connected classification layer.
Compared with the baseline CNN, ResNet provides greater representational capacity while retaining the inductive bias of local convolutional feature extraction.

\paragraph{Transformer.}

In contrast to convolution-based models, the Transformer treats each spectral point as an individual token and models the Raman spectrum as a sequence. 
Spectral intensity values are first projected into a latent embedding space, and wavenumber information is incorporated through positional encoding computed directly from the physical wavenumber coordinates rather than discrete spectral indices. 
This design preserves the correspondence between spectral features and their physical locations along the Raman axis while avoiding reliance on a predefined discretization grid. 
The embedded sequence is processed through a stack of Transformer encoder blocks, allowing the model to capture long-range dependencies and global interactions between distant spectral regions. 
A dedicated classification token is used to aggregate sequence-level information before prediction. 
Because attention is computed directly on spectral coordinates, the Transformer provides a natural framework for evaluating the benefits of the proposed wavenumber-aware representation under globally connected spectral modeling.

\subsection{Model training and evaluation}

The bacterial identification task was formulated as a multi-label classification problem,\cite{joung2026defining} where each spectrum was assigned a binary label vector indicating the presence or absence of \textit{B. atrophaeus} (BA), \textit{E. coli} (EC), and \textit{L. innocua} (LI). 
The output layer consisted of three independent logits, and models were trained using the binary cross-entropy loss with logits (BCEWithLogitsLoss).

For each bacterial species, predicted probabilities were obtained by applying a sigmoid activation function to the output logits. 
A bacterial species was considered present when the predicted probability exceeded 0.5.

Model parameters were optimized using the AdamW optimizer with a cosine learning-rate schedule. 
Early stopping based on validation loss was employed to select the final model used for testing.

Performance was evaluated using Exact Match accuracy and micro-averaged F1 score (F1$_{\mathrm{micro}}$). 
Exact Match accuracy represents the fraction of spectra for which all bacterial labels were predicted correctly,

\[
\mathrm{Exact\ Match}
=
\frac{1}{N}
\sum_{i=1}^{N}
\mathbb{I}
(\hat{\mathbf y}_i = \mathbf y_i),
\]

where \(N\) denotes the number of spectra and \(\hat{\mathbf y}_i\) and \(\mathbf y_i\) are the predicted and true label vectors, respectively.

The micro-averaged F1 score was calculated by aggregating true positives (TP), false positives (FP), and false negatives (FN) across all bacterial classes,

\[
F1_{\mathrm{micro}}
=
\frac{2TP}
     {2TP + FP + FN}.
\]

For the sliding-window analysis, the macro-averaged F1 score (F1$_{\mathrm{macro}}$) was additionally reported to evaluate performance uniformly across bacterial species,

\[
F1_{\mathrm{macro}}
=
\frac{1}{C}
\sum_{i=1}^{C}
F1_i,
\]

where \(C\) is the number of bacterial species and \(F1_i\) denotes the F1 score of class \(i\).

\subsection{Positional-encoding baselines and resampled-grid evaluation}

For the comparison in Table~\ref{tab:pe_comparison}, four Transformer variants were trained that differ only in the positional information supplied to the encoder.
The \emph{wavenumber} variant computes sinusoidal encodings directly from the physical coordinate $\omega$ (in cm$^{-1}$) of each spectral point.
The \emph{index} variant computes the same sinusoidal map from the sequence position $0,1,2,\ldots$ of each token; following standard practice for variable-length inputs, the position count restarts from zero after any windowing, so the encoding reflects the position within the observed sequence rather than within the original spectrum.
The \emph{learned} variant replaces the analytic map with a trainable embedding table over sequence positions, initialized from a scaled normal distribution and optimized jointly with the model.
The \emph{constant} variant provides no positional information; because setting the $\omega$ channel to a constant renders the positional term uninformative, this configuration coincides with the intensity-only ($I$+aug.) setting of Table~\ref{tab:pe_comparison}.
MLP-based coordinate embeddings (applied to either $\omega$ or the index) were also evaluated and are reported in the Supplementary Information.
All variants share identical architecture, data splits, synthetic-mixture generation, augmentation, and optimization settings.

For the resampled-grid evaluation (Table~\ref{tab:offset_grid}), test spectra were linearly resampled onto uniform wavenumber grids with spacings of 1, 2, and 4~cm$^{-1}$ spanning the native spectral range, producing coordinate values that do not coincide with the native, nonuniform measurement grid used during training.
For the benchmark, the same procedure was applied with spacings of 1, 1.5, 2, and 4~cm$^{-1}$; spectra longer than the fixed input length of 1{,}024 points were truncated at the high-wavenumber end.
No model was retrained for this evaluation; the identical trained models were applied directly to the resampled inputs.

\subsection{External benchmark experiments on the Ho et al.\ dataset}

The public bacteria-ID benchmark\cite{ho2019rapid} provides 60{,}000 reference spectra (30 isolates $\times$ 2{,}000 spectra, 1-s exposure), distributed with fifth-order polynomial background correction and per-spectrum min--max normalization applied; each spectrum contains 1{,}000 points spanning 381.98--1792.4~cm$^{-1}$.
We used the distributed data without additional preprocessing.
The reference set was divided into training, validation, and test subsets of 42{,}000, 9{,}000, and 9{,}000 spectra using a class-stratified random split.
The multi-label mixture formulation of the in-house experiments was applied unchanged: synthetic mixtures of one to three isolates were generated within each split, and augmentation comprised contiguous windowing to 1{,}000, 512, or 256 points (probabilities 0.4/0.3/0.3), downsampling with strides of 1--4, and additive Gaussian noise with $\sigma$ drawn from $[0, 0.05]$.
Evaluation scenarios mirror the in-house protocol (noise $\sigma=0.025$; 250-point windows for Slice; strides of 2--4 for Downsample).
The Transformer used $d_{\mathrm{model}}=256$, four encoder layers, eight attention heads, and a feed-forward dimension of 1024 (3.23M parameters), trained with AdamW (learning rate $10^{-4}$, batch size 32); CNN and ResNet baselines used base channel widths of 64.

For comparison with the published evaluation protocol, the model pretrained on the reference training split was fine-tuned on the distributed fine-tuning set (2{,}700 spectra, 2-s exposure) with a learning rate of $10^{-5}$ (one tenth of the pretraining rate, matching the ratio used by Ho et al.) and evaluated on the distributed test set (3{,}000 spectra, 2-s exposure).
Empiric-treatment groupings follow the eight antibiotic groups defined in Ref.~\citenum{ho2019rapid}.

\backmatter

\bmhead{Data availability}
The in-house bacterial Raman dataset used in this study is publicly available at \url{https://github.com/sailgroup/Wavenumber-Aware-Deep-Learning}.
The public bacteria-ID benchmark of Ho et al.\ is available at \url{https://github.com/csho33/bacteria-ID}.

\bmhead{Code availability}
The complete source code, including model implementations, training scripts, and the experiment configurations used in this study, is available at \url{https://github.com/sailgroup/Wavenumber-Aware-Deep-Learning}.
The repository also includes scripts that reproduce the central experiments of this work on the public benchmark, namely the positional-encoding comparisons, robustness evaluations, and mixture experiments.

\bmhead{Acknowledgements}
This work was supported by Korea Institute of Military Technology Planning and Advancement (KRIT) grant funded by the Korea government (DAPA) (KRIT-CT-21-034).

\bmhead{Author contributions}
J.K. performed the modelling and data analysis.
C.K. contributed to the development of the machine-learning models.
J.H., Y.K. and S.B.Y. prepared the bacterial samples and acquired the Raman spectra.
K.-I.O. and H.B.N. contributed to the discussion of the results.
K.-I.O., H.B.N., K.K., M.P. and J.F.J. conceived the study.
J.K. and J.F.J. wrote the manuscript.

\bmhead{Competing interests}
The authors declare no competing interests.

\bmhead{Supplementary information}
The online version contains Supplementary Information, which includes details of the model architectures, hyperparameter optimization and training, prediction behavior under spectral degradation, complete ablation results, additional positional-encoding results, and additional results on the public benchmark.

\bibliography{references}

\clearpage


\section*{Supplementary material}

## Section S1: Model architectures and hyperparameter optimization

To evaluate the proposed wavenumber-aware representation across different neural network architectures, three backbone models were considered: a one-dimensional convolutional neural network (CNN), a residual convolutional neural network (ResNet), and a Transformer encoder. All models received Raman spectra as a two-channel input consisting of intensity and wavenumber values. Missing spectral regions generated during slicing or downsampling experiments were padded and explicitly handled during model training and inference.

### 1.1 CNN

The CNN architecture consisted of four convolutional blocks, each composed of a one-dimensional convolution layer, batch normalization, ReLU activation, and max pooling. The number of feature channels increased progressively from $C$ to $2C$, $4C$, and $8C$, where $C$ denotes the base channel dimension. The four blocks used kernel sizes of 7, 5, 3, and 3 and max-pooling factors of 4, 4, 4, and 2. Following feature extraction, global average pooling was applied and the resulting feature vector was passed to a two-layer multilayer perceptron for classification. The global average pooling was weighted by the fraction of measured points in each pooled region, so padded positions did not contribute to the pooled features.

### 1.2 ResNet

The ResNet architecture employed an initial convolutional stem followed by four residual stages. Each residual block consisted of two one-dimensional convolution layers with batch normalization and ReLU activation connected through identity skip connections. Global average pooling was applied to the final feature maps prior to classification. The stem consisted of a strided convolution with a kernel size of 7 followed by max pooling, and the second to fourth stages halved the spectral length through strided convolutions. Depending on the hyperparameter configuration, both the network width and depth were varied. The final configuration used two residual blocks per stage, as in ResNet-18, with a base channel dimension of 64. Padded positions were excluded from the global average pooling in the same way as in the CNN.

### 1.3 Transformer

The Transformer architecture treated each spectral point as an independent token. Intensity values were first projected into a latent embedding space using a linear embedding layer. Wavenumber information was incorporated through sinusoidal positional encoding calculated directly from the physical Raman shift values rather than sequence indices. A learnable classification token (`[CLS]`) was prepended to the input sequence and processed by a stack of Transformer encoder layers employing multi-head self-attention and feed-forward networks. The final spectrum representation was obtained from the `[CLS]` token and used for bacterial classification. The linear intensity embedding was followed by layer normalization. The encoder comprised four layers with a model dimension of 256, eight attention heads, a feed-forward dimension of 1024, GELU activations, and layer normalization applied before each sublayer. Padded positions were excluded from attention through a key-padding mask. The classification head was a two-layer multilayer perceptron.

### 1.4 Hyperparameter optimization

Hyperparameters were optimized independently for each model using the validation set. The search ranges and selected values are summarized in Supplementary Table 1. The final configuration for each model was selected based on the lowest validation loss. All models were trained using the AdamW optimizer and the binary cross-entropy loss of the multi-label formulation

described in the Methods section of the main text. During the hyperparameter search, each configuration was trained for 100 epochs.

**Supplementary Table 1** Hyperparameter search ranges for each model. Selected values are shown in bold.

| Model | Hyperparameter | Search range |
|---|---|---|
| CNN | Learning rate | $5 \times 10^{-5}$, $1 \times 10^{-4}$, $\mathbf{3}\times10^{-4}$ |
| | Batch size | **32**, 64 |
| | Dropout | **0.1**, 0.2 |
| | Base channels | 32, **64** |
| ResNet | Learning rate | $5 \times 10^{-5}$, $1 \times 10^{-4}$, $\mathbf{3}\times10^{-4}$ |
| | Batch size | **32**, 64 |
| | Dropout | **0.1**, 0.2 |
| | Base channels | 32, **64** |
| Transformer | Learning rate | $5 \times 10^{-5}$, $\mathbf{1}\times10^{-4}$, $3 \times 10^{-4}$ |
| | Batch size | **16**, 32 |
| | Dropout | **0.1**, 0.2 |
| | Embedding dimension | 128, **256** |

## 1.5 Training details

All models on the in-house dataset were trained with AdamW using a weight decay of 0.01. The learning rate followed a cosine schedule without warmup and decayed to 1% of its initial value over 300 epochs. Gradients were clipped to a norm of 1.0, and mixed-precision (fp16) arithmetic was used. Each training epoch used 6,000 synthetic mixtures generated anew from the training split with an epoch-dependent random seed. Validation used a fixed set of 750 synthetic mixtures from the validation split. These mixtures received the same windowing and downsampling as the training data but no noise. Training ran for at most 300 epochs with early stopping on the validation loss (patience of 30 epochs). The checkpoint with the lowest validation loss was used for testing. All results in this work, including those on the public benchmark, were obtained from a single training run with a random seed of 42.

## Section S2: Prediction behavior under spectral degradation

To further examine the prediction behavior of the Transformer model used for the analyses in the main text, Supplementary Fig. 1 shows the row-normalized confusion matrices of the Transformer under both the unperturbed and combined degradation conditions.

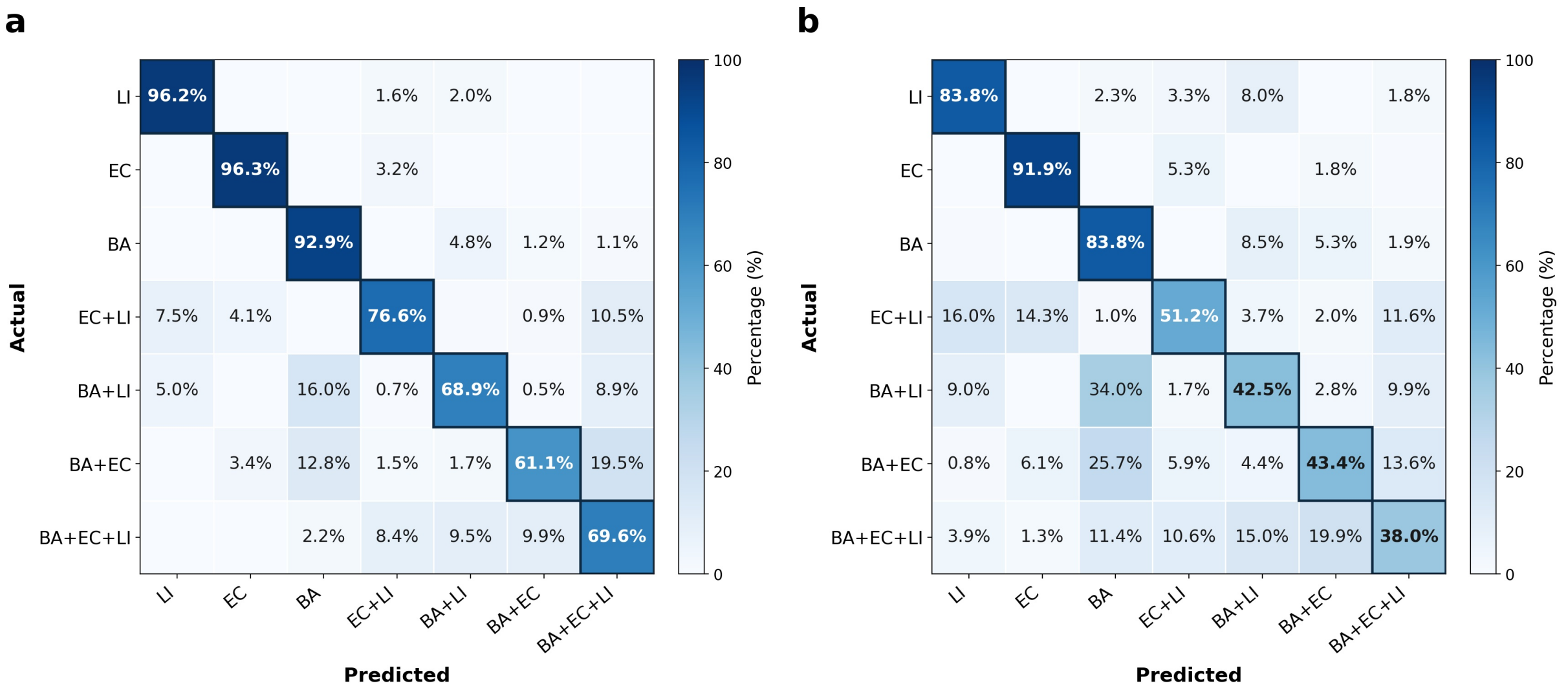


**Supplementary Fig. 1** Row-normalized confusion matrices of the Transformer model for bacterial mixture identification, where BA, EC, and LI denote *B. atrophaeus*, *E. coli*, and *L. innocua*, respectively. (a) Unperturbed condition. (b) Combined degradation condition including additive noise, partial spectral observation, and downsampling. Most prediction errors occur among mixtures with similar bacterial compositions, whereas single-species samples remain well distinguished under both conditions.

Under the unperturbed condition (Supplementary Fig. 1a), single-species samples were classified with high accuracy, achieving correct classification rates of 96.2%, 96.3%, and 92.9% for *L. innocua*, *E. coli*, and *B. atrophaeus*, respectively. In contrast, classification accuracy generally decreased as the number of bacterial species present in the spectrum increased, reflecting the increasing difficulty of resolving overlapping Raman signatures originating from multiple bacterial species. Most prediction errors occurred between mixtures sharing similar constituent species rather than between single-species and mixed-species classes, indicating that the model largely preserved the overall compositional relationships among bacterial spectra.

A similar prediction pattern was observed under the combined degradation condition (Supplementary Fig. 1b). Although classification accuracy decreased across most classes, the overall confusion structure remained largely unchanged. Single-species samples continued to exhibit relatively high classification accuracy (83.8%, 91.9%, and 83.8% for *L. innocua*, *E. coli*, and *B. atrophaeus*, respectively), whereas additional errors were primarily concentrated among multi-species mixtures, for which the correct classification rates decreased from 61–77% to 38–51%. Notably, degradation increased confusion between compositionally related classes, in which one constituent of a mixture is missed (e.g., BA+LI predicted as BA) or a mixture is confused with the three-species class, rather than introducing systematic confusion between unrelated classes.

Together with the results presented in the main text, these confusion matrices indicate that spectral degradation primarily reduces the confidence with which overlapping bacterial signatures are resolved, without fundamentally altering the learned relationships among bacterial classes. This observation further supports the robustness of the proposed wavenumber-aware representation under realistic Raman acquisition conditions.

## Section S3: Ablation study

To quantify the individual contributions of the proposed wavenumber-aware representation and data augmentation strategy, an ablation study was conducted using CNN, ResNet, and Transformer architectures. Four training configurations were evaluated: (i) intensity-only spectra ($I$ only), (ii) intensity and wavenumber inputs ($I + \omega$), (iii) intensity-only spectra with augmentation ($I$+aug.), and (iv) wavenumber-aware spectra with augmentation ($I + \omega$+aug.). Model performance was evaluated under five testing scenarios: unperturbed spectra, noisy spectra, sliced spectra, downsampled spectra, and spectra subjected to both slicing and downsampling.

Supplementary Tables 2–6 summarize the Exact Match and micro-averaged F1 scores obtained for all architectures and evaluation conditions.

Across all architectures, the combination of wavenumber information and augmentation consistently produced the highest robustness under spectral degradation. While augmentation alone substantially improved performance under noisy, sliced, and downsampled conditions, incorporating explicit wavenumber information provided additional gains, particularly when spectral coverage or sampling density differed from those observed during training.

For the CNN and ResNet models, augmentation was the dominant factor affecting robustness. Under the Noise+Slice+Downsample condition, the F1 score increased from 48.65% to 83.78% for CNN and from 47.70% to 82.99% for ResNet when augmentation was introduced. The addition of wavenumber information further improved performance, yielding the best overall results for both architectures.

In contrast, the Transformer architecture exhibited a substantially stronger dependence on wavenumber information. Using intensity values alone resulted in an unperturbed-spectrum F1 score of only 84.75%, whereas incorporating wavenumber coordinates increased the score to 95.71%. Similar improvements were observed under noise and downsampling, whereas under scenarios involving partial observation the benefit of the wavenumber channel without augmentation largely disappeared, consistent with the main-text finding that explicit coordinates confer robustness only in combination with augmentation. This result is consistent with the Transformer architecture, in which positional information directly determines how spectral points are interpreted by the self-attention mechanism. Explicit physical coordinates therefore provide a more meaningful positional reference than fixed-grid indices when spectral resolution or spectral coverage varies.

The strongest performance was consistently obtained using the full $I + \omega$+aug. configuration. For the Transformer model, this configuration achieved F1 scores of 93.78%, 92.98%, 87.65%, 91.31%, and 86.25% under the Unperturbed, Noise, Noise+Slice, Noise+Downsample, and Noise+Slice+Downsample conditions, respectively (Table 1 in the main text; the Transformer $F1_{micro}$ values for the other configurations are given in Table 3 of the main text). These results demonstrate that data augmentation and wavenumber-aware representation provide complementary benefits, with augmentation improving invariance to spectral degradation and explicit wavenumber information preserving the physical meaning of spectral observations.

**Supplementary Table 2** CNN ablation study (Exact Match, %).

| Scenario | $I$ only | $I + \omega$ | $I$+aug. | $I + \omega$+aug. |
|---|---|---|---|---|
| Unperturbed | 84.05 | 85.23 | 76.22 | 79.95 |
| Noise | 62.54 | 66.63 | 75.22 | 78.74 |
| Noise+Slice | 17.89 | 18.26 | 54.69 | 59.98 |
| Noise+Downsample | 23.26 | 28.81 | 71.90 | 73.82 |
| Noise+Slice+Downsample | 15.08 | 11.95 | 55.70 | 59.05 |

**Supplementary Table 3** ResNet ablation study (Exact Match, %).

| Scenario | $I$ only | $I + \omega$ | $I$+aug. | $I + \omega$+aug. |
|---|---|---|---|---|
| Unperturbed | 81.82 | 82.82 | 73.34 | 81.48 |
| Noise | 56.12 | 65.92 | 71.46 | 79.48 |
| Noise+Slice | 18.83 | 17.43 | 55.16 | 64.02 |
| Noise+Downsample | 22.95 | 22.66 | 70.58 | 74.75 |
| Noise+Slice+Downsample | 15.67 | 11.26 | 54.37 | 61.00 |

**Supplementary Table 4** Transformer ablation study (Exact Match, %).

| Scenario | $I$ only | $I + \omega$ | $I$+aug. | $I + \omega$+aug. |
|---|---|---|---|---|
| Unperturbed | 56.70 | 85.99 | 59.90 | 80.02 |
| Noise | 52.39 | 71.67 | 58.97 | 77.88 |
| Noise+Slice | 18.75 | 20.53 | 40.97 | 64.94 |
| Noise+Downsample | 47.98 | 59.35 | 57.06 | 73.59 |
| Noise+Slice+Downsample | 22.42 | 20.98 | 40.89 | 61.61 |

**Supplementary Table 5** CNN ablation study ($F1_{micro}$, %).

| Scenario | $I$ only | $I + \omega$ | $I$+aug. | $I + \omega$+aug. |
|---|---|---|---|---|
| Unperturbed | 95.05 | 95.40 | 92.49 | 93.68 |
| Noise | 84.70 | 86.47 | 91.97 | 93.18 |
| Noise+Slice | 50.28 | 53.66 | 83.37 | 85.61 |
| Noise+Downsample | 63.10 | 66.27 | 90.72 | 91.33 |
| Noise+Slice+Downsample | 48.65 | 48.58 | 83.78 | 85.25 |

**Supplementary Table 6** ResNet ablation study ($F1_{micro}$, %).

| Scenario | $I$ only | $I + \omega$ | $I$+aug. | $I + \omega$+aug. |
|---|---|---|---|---|
| Unperturbed | 94.42 | 94.59 | 91.61 | 94.27 |
| Noise | 82.07 | 86.38 | 90.51 | 93.54 |
| Noise+Slice | 51.68 | 50.89 | 83.33 | 87.52 |
| Noise+Downsample | 59.32 | 55.70 | 90.26 | 91.78 |
| Noise+Slice+Downsample | 47.70 | 41.47 | 82.99 | 86.06 |

## Section S4: Additional positional-encoding results

This section reports the Exact Match counterparts of the positional-encoding comparison presented in the main text, the MLP-based coordinate-embedding baselines, and the complete resampled-grid results for all five evaluation scenarios.

**Supplementary Table 7** Comparison of positional-encoding schemes under identical data, augmentation, and training conditions (Transformer, Exact Match, %). All configurations use augmented training.

| Scenario | Constant | Index | Learned | Wavenumber (proposed) |
|---|---|---|---|---|
| Unperturbed | 59.90 | 78.85 | 66.13 | **80.02** |
| Noise | 58.97 | 77.53 | 64.68 | **77.88** |
| Noise+Slice | 40.97 | 51.47 | 41.91 | **64.94** |
| Noise+Downsample | 57.06 | 71.96 | 62.53 | **73.59** |
| Noise+Slice+Downsample | 40.89 | 51.20 | 42.60 | **61.61** |

In addition to the analytic and learned positional encodings discussed in the main text, coordinate embeddings computed by a multilayer perceptron (MLP) were evaluated, applied either to the physical wavenumber, MLP($\omega$), or to the sequence index, MLP(index). Both variants underperform the sinusoidal wavenumber encoding under all scenarios, and the index-based variant additionally degrades under partial observation, consistent with the behavior of the sinusoidal index encoding.

**Supplementary Table 8** MLP-based coordinate embeddings evaluated on the native grid and on uniform resampled grids with 1, 2, and 4 $cm^{-1}$ spacing (Transformer, augmented training).

| | MLP($\omega$) | | | | MLP(index) | | | |
|---|---|---|---|---|---|---|---|---|
| Scenario | Native | 1 | 2 | 4 | Native | 1 | 2 | 4 |
| $F1_{micro}$ (%) | | | | | | | | |
| Unperturbed | 91.14 | 90.99 | 90.83 | 90.40 | 90.43 | 89.91 | 89.94 | 89.22 |
| Noise | 90.69 | 90.82 | 90.39 | 89.70 | 90.01 | 90.01 | 89.71 | 88.87 |
| Noise+Slice | 84.71 | 84.77 | 84.48 | 83.87 | 82.65 | 82.60 | 82.52 | 81.82 |
| Noise+Downsample | 89.73 | 90.02 | 89.73 | 89.37 | 89.25 | 88.72 | 89.04 | 88.63 |
| Noise+Slice+Downsample | 84.19 | 84.08 | 84.14 | 83.83 | 82.16 | 81.85 | 82.15 | 81.59 |
| Exact Match (%) | | | | | | | | |
| Unperturbed | 72.44 | 71.33 | 71.08 | 69.95 | 71.34 | 68.53 | 69.14 | 67.00 |
| Noise | 71.64 | 71.52 | 70.36 | 68.68 | 70.66 | 69.26 | 69.08 | 66.45 |
| Noise+Slice | 56.96 | 56.95 | 56.03 | 54.76 | 53.26 | 52.81 | 52.36 | 50.52 |
| Noise+Downsample | 69.16 | 69.20 | 68.46 | 68.03 | 68.21 | 65.75 | 67.23 | 65.77 |
| Noise+Slice+Downsample | 55.96 | 55.27 | 55.29 | 54.80 | 51.91 | 50.80 | 51.28 | 50.34 |

**Supplementary Table 9** Complete resampled-grid results for all five evaluation scenarios (Transformer, $F1_{micro}$, %). Evaluation spectra were resampled onto uniform grids with the indicated spacing; "Native" denotes the original measurement grid.

| Scenario | Encoding | Native | 1 $cm^{-1}$ | 2 $cm^{-1}$ | 4 $cm^{-1}$ |
|---|---|---|---|---|---|
| Unperturbed | Wavenumber | 93.78 | 93.74 | 93.68 | 92.44 |
| | Index | 93.37 | 72.50 | 78.38 | 82.37 |
| | Constant | 86.39 | 85.80 | 85.64 | 84.80 |
| Noise | Wavenumber | 92.98 | 93.17 | 92.79 | 91.15 |
| | Index | 92.96 | 72.09 | 77.95 | 82.38 |
| | Constant | 86.03 | 85.50 | 85.31 | 84.58 |
| Noise+Slice | Wavenumber | 87.65 | 88.01 | 87.47 | 86.31 |
| | Index | 81.69 | 79.37 | 79.80 | 81.37 |
| | Constant | 77.72 | 77.84 | 78.09 | 77.79 |
| Noise+Downsample | Wavenumber | 91.31 | 91.02 | 91.23 | 90.78 |
| | Index | 90.72 | 72.15 | 78.23 | 82.21 |
| | Constant | 85.15 | 85.06 | 84.82 | 84.23 |
| Noise+Slice+Downsample | Wavenumber | 86.25 | 86.69 | 86.52 | 85.84 |
| | Index | 81.77 | 79.04 | 79.34 | 80.94 |
| | Constant | 77.79 | 77.34 | 77.73 | 77.33 |

## Section S5: External benchmark: additional results and protocol comparison

This section provides additional results for the external validation on the public benchmark of Ho et al.,[1] including architecture-level comparisons, synthetic-mixture results, the comparison with the published evaluation protocol, and the exposure-shift analysis.

The synthetic-mixture experiments on the benchmark use the multi-label formulation of the main text. The single-spectrum experiments differ in one respect. Because each benchmark spectrum belongs to exactly one of 30 isolates, these models were trained as single-label 30-class classifiers with a softmax output and the cross-entropy loss, and performance is reported as Top-1 accuracy. All other settings, including the input representation, augmentation, architecture, and optimizer, follow the mixture experiments.

**Supplementary Table 10** Architecture comparison on the Ho et al. benchmark for single spectra (30-class Top-1 and empiric-treatment 8-group accuracy, %; $I + \omega$+aug.).

| | CNN | | ResNet | | Transformer | |
|---|---|---|---|---|---|---|
| Scenario | Top-1 | Group | Top-1 | Group | Top-1 | Group |
| Unperturbed | 94.41 | 99.13 | 95.49 | 99.43 | 95.90 | 99.59 |
| Noise only | 94.18 | 99.08 | 95.11 | 99.28 | 95.53 | 99.58 |
| Noise+Slice | 65.22 | 82.99 | 70.23 | 86.92 | 73.91 | 89.48 |
| Noise+Downsample | 83.83 | 94.90 | 85.13 | 95.70 | 86.57 | 96.39 |
| Noise+Slice+Downsample | 50.93 | 71.91 | 55.03 | 74.88 | 58.36 | 77.57 |

**Supplementary Table 11** Architecture comparison on the Ho et al. benchmark for synthetic mixtures of up to three isolates (Exact Match and $F1_{micro}$, %; $I + \omega$+aug.).

| | CNN | | ResNet | | Transformer | |
|---|---|---|---|---|---|---|
| Scenario | Exact | $F1_{micro}$ | Exact | $F1_{micro}$ | Exact | $F1_{micro}$ |
| Unperturbed | 44.87 | 71.84 | 49.53 | 75.67 | 57.17 | 80.08 |
| Noise only | 44.34 | 71.41 | 48.63 | 74.86 | 56.21 | 79.22 |
| Noise+Slice | 17.97 | 36.71 | 24.53 | 45.68 | 29.88 | 51.69 |
| Noise+Downsample | 36.36 | 61.63 | 39.57 | 64.57 | 43.98 | 68.13 |
| Noise+Slice+Downsample | 12.24 | 26.99 | 15.07 | 31.61 | 19.14 | 37.79 |

### 5.1 Comparison with the published evaluation protocol

The published evaluation of Ho et al. fine-tunes the pretrained model on the distributed 2-s fine-tuning set and evaluates it on the test set distributed with the benchmark, whereas the reference training data were acquired at 1-s exposure. For fine-tuning, the distributed fine-tuning set was split into training and validation subsets, and the pretrained model was trained for at most 30 epochs with a learning rate of $10^{-5}$ and a batch size of 32. The same augmentation as in pretraining was applied, and early stopping on the validation loss used a patience of 5 epochs. Supplementary Table 12 separates the two evaluation settings. On a random stratified split of the 1-s reference set, an in-domain setting, the wavenumber-aware Transformer attains 95.90% Top-1 and 99.59% group accuracy. Replicating the published protocol, the model reaches 83.00% Top-1 and 96.47% group accuracy, comparable to the reported 82.2±0.3% and 97.0±0.3%.

### 5.2 Exposure-shift analysis

The transfer from the reference set to the distributed test set involves an acquisition-state shift: the exposure time was increased from 1 s to 2 s to compensate for a documented degradation

**Supplementary Table 12** Protocol-separated comparison on the Ho et al. benchmark (30-class Top-1 and empiric-treatment 8-group accuracy, %). Literature values are as reported in Ref. 1.

| | 30-class | 8-group |
|---|---|---|
| *Reference test, 1-s exposure (random stratified split)* | | |
| This work ($I + \omega$+aug.) | 95.90 | 99.59 |
| *Distributed test set, 2-s exposure* | | |
| This work ($I + \omega$+aug.), fine-tuned | 83.00 | 96.47 |
| CNN, fine-tuned (Ho et al.) | 82.2±0.3 | 97.0±0.3 |
| Logistic regression (Ho et al.) | 75.7 | 93.3 |
| SVM (Ho et al.) | 74.9 | 92.2 |

of optical-system efficiency,[1] so the test spectra are systematically less noisy than the training spectra. This direction is opposite to the noise-oriented augmentation used in this work, which trains the model to tolerate additional noise rather than reduced noise. Supplementary Table 13 reports the zero-shot results on the unperturbed scenario, obtained by evaluating models trained solely on 1-s spectra without any fine-tuning. The intensity-only configurations collapse to around 10% Top-1 accuracy (8.90% and 10.00%), whereas adding the wavenumber channel raises zero-shot accuracy to 41.60%, and augmentation provides a further gain to 41.97% (62.80% at the treatment-group level). In other words, most of the structure that transfers across this exposure shift is carried by the explicit wavenumber coordinates, and a model trained only on 1-s spectra already recovers a substantial part of the 2-s behavior without seeing a single 2-s spectrum. Fully zero-shot transfer across acquisition states nevertheless remains incomplete: fine-tuning on the 2,700 distributed 2-s spectra raises the accuracy to 83.00% (96.47% at the group level), so a small amount of target-condition data is still required to reach benchmark-level performance.

**Supplementary Table 13** Zero-shot evaluation on the distributed 2-s test set of the Ho et al. benchmark without fine-tuning (Transformer trained on 1-s reference spectra; unperturbed scenario). 30-class Top-1 and empiric-treatment group accuracy (%). The fine-tuned result is shown for reference.

| Input configuration | Top-1 | Group |
|---|---|---|
| $I$ | 8.90 | 25.50 |
| $I$+aug. | 10.00 | 27.90 |
| $I + \omega$ | 41.60 | 61.33 |
| $I + \omega$+aug. | 41.97 | 62.80 |
| $I + \omega$+aug., after fine-tuning | 83.00 | 96.47 |

## 5.3 Input-representation ablation and class-wise predictions

Supplementary Table 14 reports the complete input-representation ablation on the benchmark, including the configurations trained without augmentation; the augmented configurations are also shown in Table 2 of the main text. Supplementary Fig. 2 shows the 30-isolate confusion matrix of the wavenumber-aware Transformer; misclassifications are concentrated within groups of closely related isolates, mirroring the structure reported by Ho et al.

**Supplementary Table 14** Input-representation ablation on the Ho et al. benchmark. Single-spectrum 30-class Top-1 accuracy and synthetic-mixture Exact Match accuracy (%, Transformer). idxPE denotes index-based sinusoidal positional encoding and lrnPE learned positional embeddings.

| Task | Scenario | $I$ | $I+\omega$ | $I$ +aug. | $I$+idxPE +aug. | $I$+lrnPE +aug. | $I+\omega$ +aug. |
|---|---|---|---|---|---|---|---|
| Single spectra (Top-1) | Unperturbed | 28.10 | 94.30 | 27.28 | 93.96 | 92.54 | **95.90** |
| | Noise+Slice | 8.01 | 17.14 | 19.43 | 42.82 | 24.81 | **73.91** |
| | Noise+Slice+Downsample | 7.19 | 16.04 | 18.51 | 49.30 | 45.19 | **58.36** |
| Mixtures (Exact Match) | Unperturbed | 7.03 | **57.60** | 6.19 | 52.57 | 49.22 | 57.17 |
| | Noise+Slice | 1.56 | 3.23 | 3.48 | 8.63 | 3.66 | **29.88** |
| | Noise+Slice+Downsample | 1.50 | 3.23 | 2.83 | 13.64 | 11.70 | **19.14** |

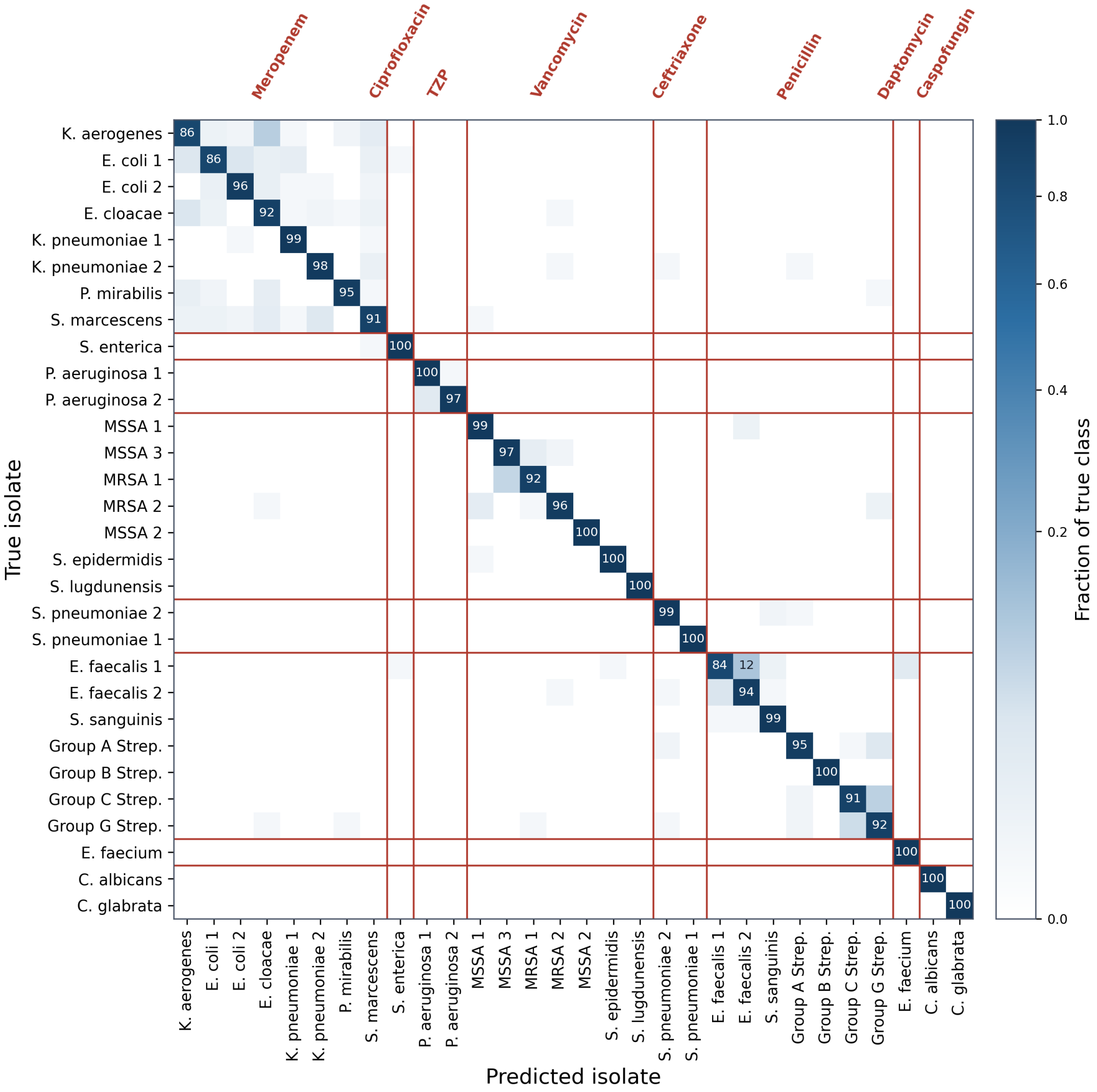


**Supplementary Fig. 2** Row-normalized 30-isolate confusion matrix on the Ho et al. benchmark (Transformer, $I + \omega$+aug., unperturbed condition). Red boxes indicate empiric-treatment groupings; misclassifications are concentrated within groups of closely related isolates.

## 5.4 Resampled-grid evaluation on the benchmark

Supplementary Table 15 repeats the resampled-grid evaluation of the main text on the benchmark. Benchmark spectra share a single native grid with a spacing of about 1.41 $cm^{-1}$. Evaluation spectra of synthetic mixtures were therefore resampled onto uniform grids of 1, 1.5, 2, and 4 $cm^{-1}$, and the models of Table 2 of the main text were evaluated without retraining. The 1 $cm^{-1}$ grid yields about 1,410 points, which exceeds the fixed input length of 1,024 points. In the scenarios without windowing, these spectra were therefore truncated at the high-wavenumber end, and the corresponding values are marked.

**Supplementary Table 15** Resampled-grid evaluation on the Ho et al. benchmark (synthetic mixtures, Transformer, augmented training, %). Evaluation spectra were resampled onto uniform grids with the indicated spacing ($cm^{-1}$); "Native" denotes the original measurement grid (about 1.41 $cm^{-1}$).

| | Wavenumber (proposed) | | | | | Index | | | | |
|---|---|---|---|---|---|---|---|---|---|---|
| Scenario | Native | 1 | 1.5 | 2 | 4 | Native | 1 | 1.5 | 2 | 4 |
| $F1_{micro}$ (%) | | | | | | | | | | |
| Unperturbed | 80.08 | 73.22[a] | 79.13 | 78.97 | 73.06 | 76.63 | 7.51[a] | 18.90 | 23.35 | 24.86 |
| Noise | 79.22 | 73.24[a] | 78.60 | 78.32 | 72.03 | 76.19 | 7.50[a] | 18.71 | 23.55 | 24.43 |
| Noise+Slice | 51.69 | 51.89 | 51.00 | 49.35 | 40.21 | 20.04 | 13.24 | 22.48 | 30.94 | 30.87 |
| Noise+Downsample | 68.13 | 59.36[a] | 65.97 | 66.76 | 67.27 | 65.15 | 7.52[a] | 16.71 | 21.74 | 24.29 |
| Noise+Slice+Downsample | 37.79 | 39.45 | 39.95 | 40.11 | 37.27 | 30.20 | 10.89 | 16.67 | 22.01 | 27.59 |
| Exact Match (%) | | | | | | | | | | |
| Unperturbed | 57.17 | 49.08[a] | 56.08 | 55.53 | 49.02 | 52.57 | 3.16[a] | 6.96 | 8.19 | 11.42 |
| Noise | 56.21 | 49.37[a] | 55.46 | 54.77 | 47.77 | 52.09 | 3.13[a] | 6.71 | 8.30 | 11.38 |
| Noise+Slice | 29.88 | 29.79 | 29.36 | 27.36 | 20.30 | 8.63 | 4.56 | 9.72 | 14.81 | 14.06 |
| Noise+Downsample | 43.98 | 35.32[a] | 41.40 | 42.60 | 42.91 | 40.58 | 2.71[a] | 5.93 | 7.53 | 10.69 |
| Noise+Slice+Downsample | 19.14 | 20.37 | 20.90 | 21.03 | 18.66 | 13.64 | 3.63 | 6.36 | 9.58 | 12.36 |

[a] The resampled spectrum exceeds the fixed input length of 1,024 points and was truncated at the high-wavenumber end.

## Section S6: Dataset composition

Supplementary Table 16 summarizes the number of in-house spectra of each species assigned to the training, validation, and test splits.

**Supplementary Table 16** Dataset composition and train/validation/test split.

| Species | Train | Validation | Test | Total |
|---|---|---|---|---|
| *B. atrophaeus* | 2000 | 250 | 250 | 2500 |
| *E. coli* | 1600 | 200 | 200 | 2000 |
| *L. innocua* | 2400 | 300 | 300 | 3000 |
| Total | 6000 | 750 | 750 | 7500 |



\end{document}